\documentclass[letterpaper, 10 pt, conference, twoside]{ieeeconf}  

\IEEEoverridecommandlockouts                              

\usepackage{amsmath} 
\usepackage{amssymb}  
\usepackage{graphicx}
\usepackage{dsfont}
\usepackage{algorithmic}
\usepackage{booktabs}
\usepackage{bm}
\usepackage{tabularx}
\usepackage{multicol}
\usepackage{multirow}
\usepackage{dblfloatfix}
\usepackage{subfig}
\usepackage{balance}
\usepackage{xcolor}
\newcommand{\ra}[1]{\renewcommand{\arraystretch}{#1}}

\usepackage[font=footnotesize,labelfont=bf]{caption}

\DeclareMathOperator*{\argmin}{argmin}

\usepackage{algorithm}

\makeatletter
\newcommand\fs@spaceruled{\def\@fs@cfont{\bfseries}\let\@fs@capt\floatc@ruled
  \def\@fs@pre{\vspace{.5\baselineskip}\hrule height.8pt depth0pt \kern2pt}%
  \def\@fs@post{\kern2pt\hrule\relax}%
  \def\@fs@mid{\kern2pt\hrule\kern2pt}%
  \let\@fs@iftopcapt\iftrue}
\makeatother

\newcommand{\updated}[1]{\textcolor{black}{#1}}

\usepackage{tikz}
\usepackage{textcomp}
\usepackage{hyperref}
\usepackage{lipsum}

\newcommand\copyrighttext{%
  \footnotesize \textcopyright 2022 IEEE. Personal use of this material is permitted.
  Permission from IEEE must be obtained for all other uses, in any current or future
  media, including reprinting/republishing this material for advertising or promotional
  purposes, creating new collective works, for resale or redistribution to servers or
  lists, or reuse of any copyrighted component of this work in other works.}
\newcommand\copyrightnotice{%
\begin{tikzpicture}[remember picture,overlay]
\node[anchor=south,yshift=10pt] at (current page.south) {\fbox{\parbox{\dimexpr\textwidth-\fboxsep-\fboxrule\relax}{\copyrighttext}}};
\end{tikzpicture}%
}

\usepackage{fancyhdr}
\fancypagestyle{firststyle}
{
   \fancyhf{}
   \fancyhead[R]{\thepage}
   \fancyhead[L]{IEEE ROBOTICS AND AUTOMATION LETTERS. PREPRINT VERSION. ACCEPTED JAN 2022}
}

\title{\LARGE \bf
Configuration Space Decomposition for Scalable Proxy Collision Checking in Robot Planning and Control}

\author{Mrinal Verghese$^{1}$, Nikhil Das$^{1}$, Yuheng Zhi$^{1}$, and Michael Yip$^{1}$

\thanks{Manuscript received: September 9, 2021; Revised:
December 18, 2021; Accepted: January 9, 2022. This paper was recommended for publication by
Editor Hanna Kurniawati upon evaluation of the Associate Editor and Reviewers’
comments. $^{1}$Mrinal Verghese, Nikhil Das, Yuheng Zhi, and Michael Yip are with the department of Electrical and Computer Engineering, University of California San Diego, La Jolla, California, USA \tt\small \{mtverghe, nrdas, yzhi, yip\}@ucsd.edu. 
\rm\footnotesize Digital Object Identifier (DOI): see top of this page.
}
}

\begin{document}

\maketitle
\copyrightnotice

\thispagestyle{firststyle}

\begin{abstract}

Real-time robot motion planning in complex high-dimensional environments remains an open problem. Motion planning algorithms, and their underlying collision checkers, are crucial to any robot control stack. Collision checking takes up a large portion of the computational time in robot motion planning. Existing collision checkers make trade-offs between speed and accuracy and scale poorly to high-dimensional, complex environments. We present a novel space decomposition method using K-Means clustering in the Forward Kinematics space to accelerate proxy collision checking. We train individual configuration space models using Fastron, \updated{a kernel perceptron algorithm}, on these decomposed subspaces, yielding compact yet highly accurate models that can be queried rapidly and scale better to more complex environments. We demonstrate this new method, called Decomposed Fast Perceptron (D-Fastron), on the 7-DOF Baxter robot producing on average 29$\times$ faster collision checks and up to 9.8$\times$ faster motion planning compared to state-of-the-art geometric collision checkers. 

\end{abstract}

\section{Introduction}
Motion planning is a crucial part of robot control. Efficiently planning in higher dimensional environments with dynamic obstacles remains an open problem \cite{MP Review}. Motion planning requires a robot to find a sequence of collision-free states from a start state to a goal state in the workspace. Planning is often done in the configuration space (C-space) of the robot, where each dimension represents a different controllable joint of the robot. Points in this space represent unique robot configurations, and working in this space can make checking for collisions and executing motion plans much easier \cite{C-Space}. However, planning in the C-space often greatly increases the dimensionality of the planning problem and necessitates more efficient planning algorithms to search the space in a reasonable amount of time.

Many approaches exist to improve the efficiency of motion planning algorithms. \updated{For efficient planners,} collision checking in particular can take up to 90$\%$ of motion planning time \cite{MP Review}\updated{\cite{Collision Checking vs Nearest Neighbors}} and is a good target for improving efficiency. Geometric collision checkers are often used to decide the collision status of a point in the C-space. While these collision checkers have near-perfect accuracy, they can be slow, which has led to a rise in the popularity of proxy collision checkers \cite{Fastron Motion Planning}\cite{FastronFK}\cite{Pan SVM}\cite{Pan KNN}\cite{Huh1}\cite{Huh2}. These methods utilize sampled, labeled datasets and parametric or non-parametric models to make predictions about a given configuration's collision status. These models can be queried an order of magnitude faster than geometric methods and can give very accurate collision predictions. 

\begin{figure}[tb] 
    \centering
  \subfloat[\label{1a}]{%
       \includegraphics[width=0.49\linewidth, trim={1.5cm 3cm 1.5cm 1cm}, clip]{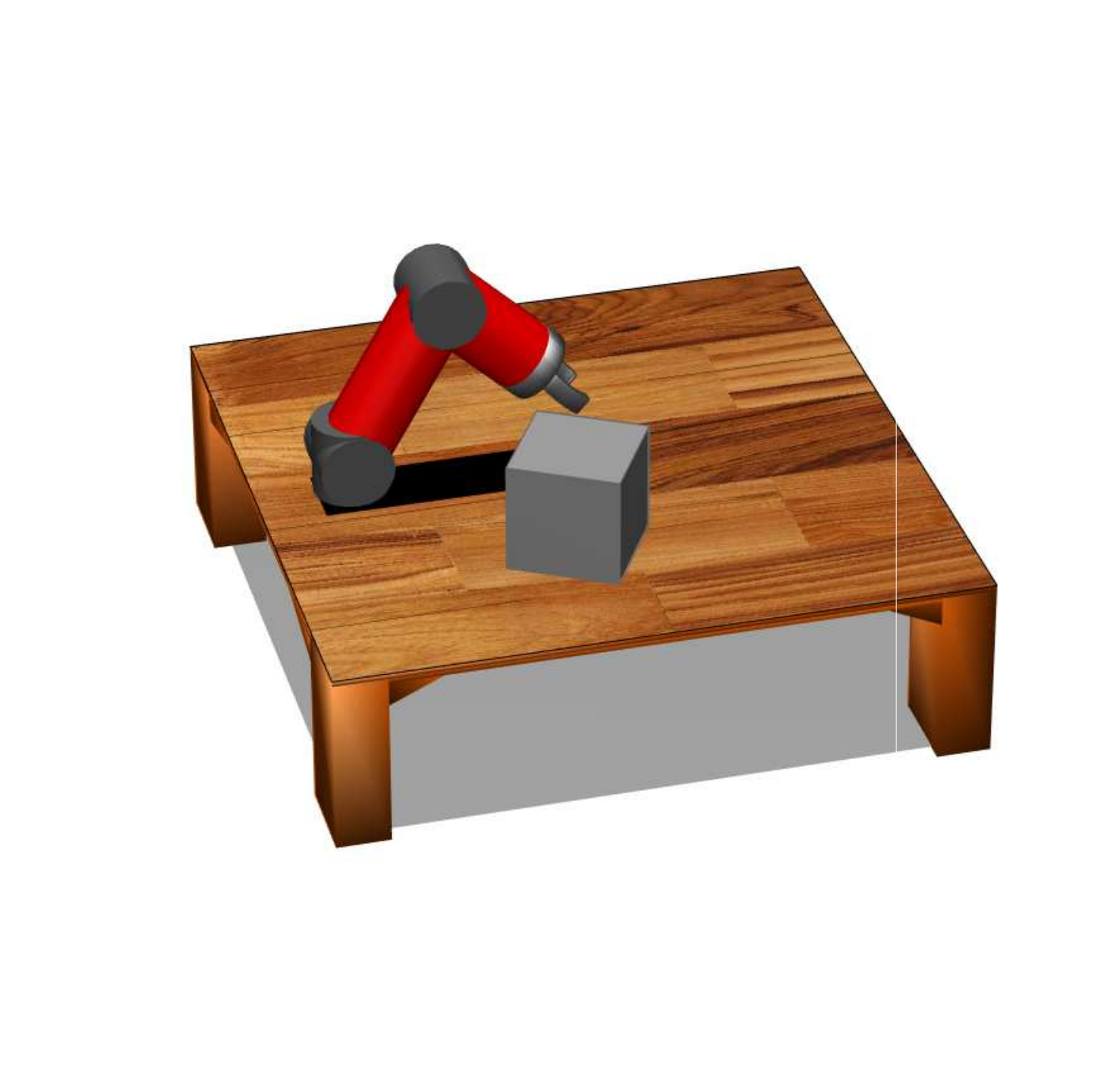}}
  \subfloat[\label{1b}]{%
        \includegraphics[width=0.49\linewidth, trim={0cm 1cm 1cm 1.5cm}, clip]{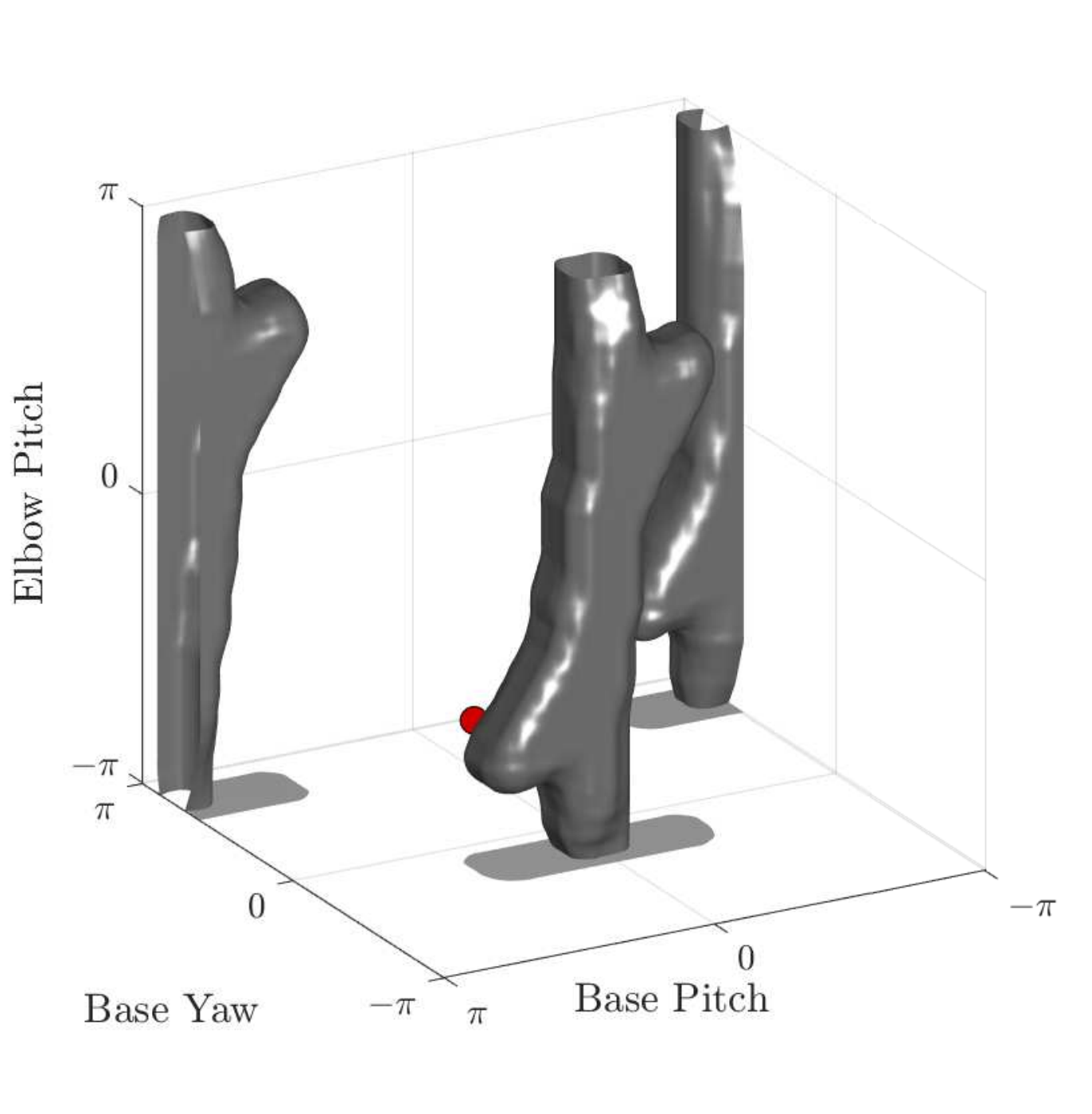}}\\
  \subfloat[\label{1c}]{%
        \includegraphics[width=0.49\linewidth, trim={0cm 1cm 0.5cm 1.5cm}, clip]{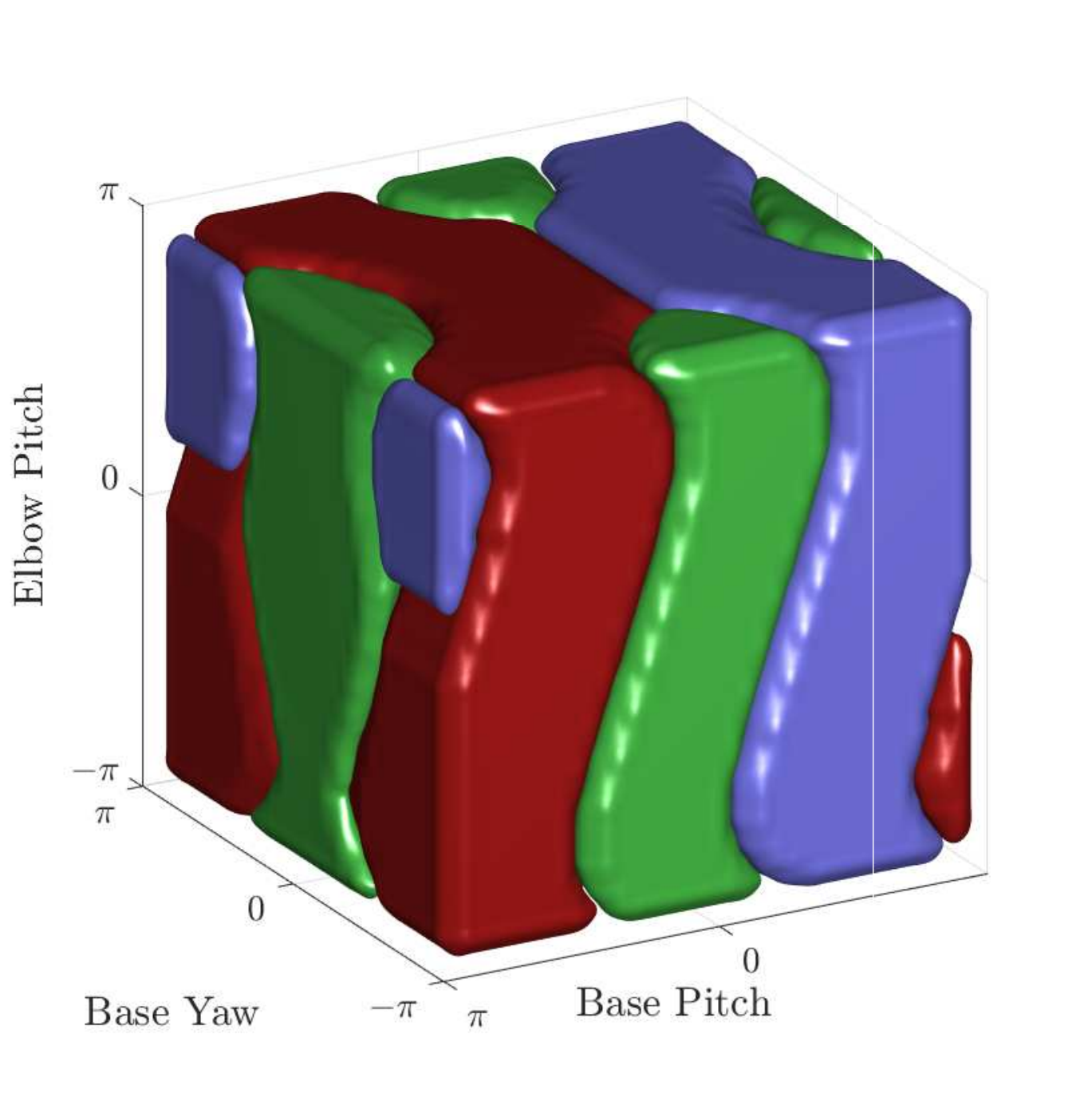}}
  \subfloat[\label{1d}]{%
        \includegraphics[width=0.49\linewidth, trim={0cm 1cm 1cm 1.5cm}, clip]{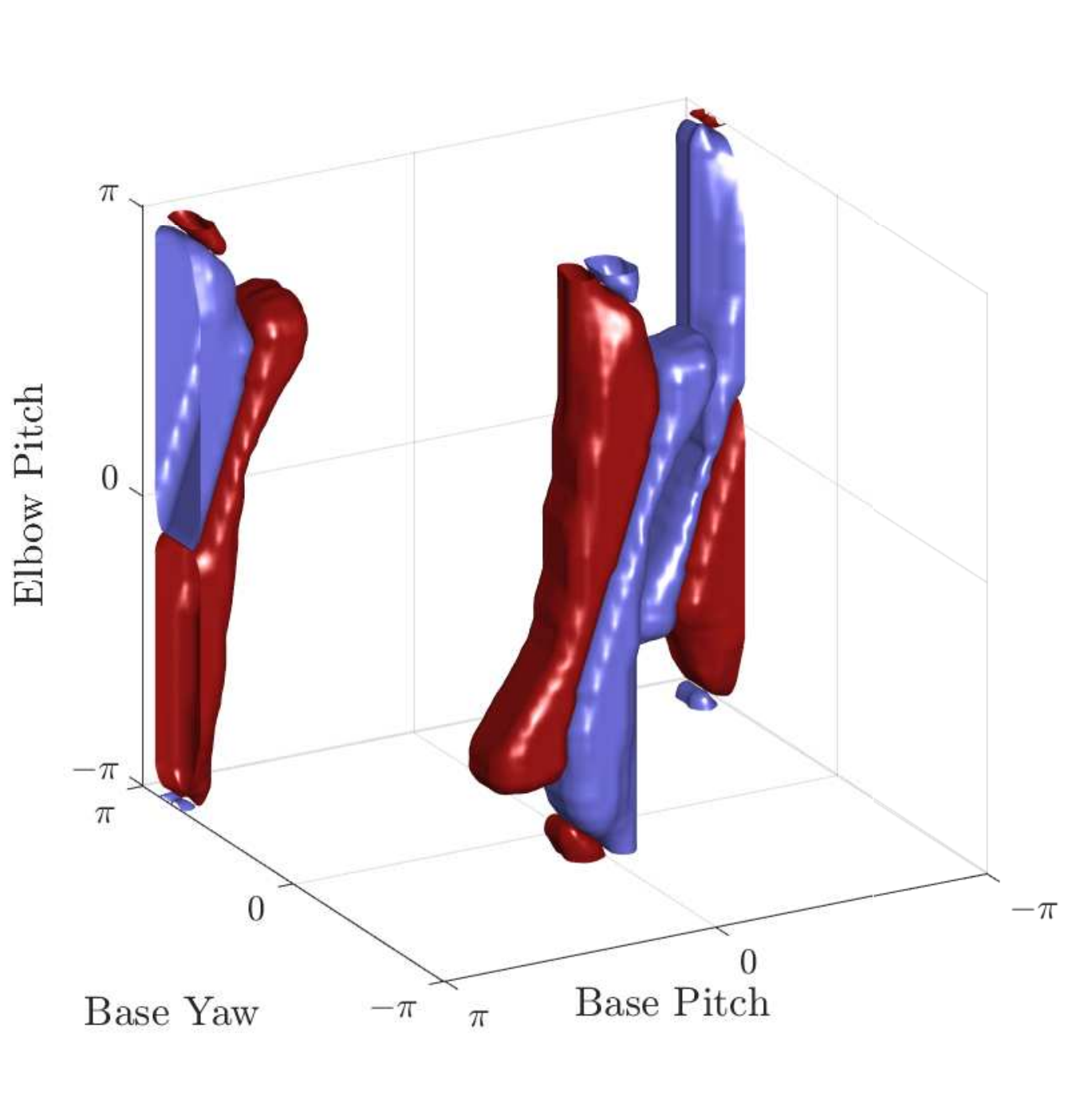}}

  \caption{This paper presents the Decomposed Fast Perceptron (D-Fastron), which decomposes robot configuration spaces for scalable collision checking. (a) A three-link robot arm with an obstacle in its workspace. (b) The configuration space representation of the same robot and obstacle. The robot configuration is represented by a red dot. (c) A decomposition of the configuration space by our method into three subspaces, color-coded with red, green, and blue. For a clearer view, see Figure \ref{fig:decomp}. (d) The configuration space obstacle partitioned according to which subspace it lies in. It is more efficient to model the resulting subspaces individually.}
  \label{fig:CSpace} 
  \vspace{-5mm}
\end{figure}


To be successful, proxy collision checkers need to scale well to high-dimensional spaces and have correct models. While the initial process of searching the space and constructing a plan can be done using the proxy checker, the plan later needs to be verified for integrity using an exact geometric collision checker and then repaired if needed. As such, proxy collision checkers rely on having high correctness to minimize the likelihood of false negatives that would lead to incorrect trajectories and excessive path repair. To maintain correctness, the complexity of the proxy model often increases with the dimensionality and complexity of the C-space, resulting in potential problems with scaling. We seek to identify strategies for maintaining accuracy and coverage of high-dimensional, complex spaces while keeping model sizes low to optimize collision checking times and model correctness. 

\subsection*{\normalfont{\textbf{\updated{Contributions}}}}
In this work, we describe a scalable, non-parametric decomposition strategy for improved proxy-based collision checking. We propose a method to intelligently decompose the robot configuration space into smaller subspaces and train independent, lightweight proxy detectors per subspace. This approach, which we call Decomposed Fast Perceptron, or D-Fastron, leads to a reduction in the complexity of each model, enabling faster queries anywhere in decomposed C-space. The method combines a K-Means decomposition method with a \updated{forward kinematics-based transform} to provide a unique configuration space decomposition for each robot. \updated{An example for a 3-DOF robot is shown in figure \ref{fig:CSpace}.} We show that D-Fastron provides substantial improvements by scalably modeling the configuration space map of collision regions and \updated{performs better} than other space decomposition methods as measured by the lower complexity of the trained models. Training the set of decomposed models in individual subspaces yields models that scale better with environment complexity, are on average 2.46$\times$ faster than a global proxy collision checking model and 29$\times$ faster than geometric models, and yield 38\%-89\% faster motion planning (on a 7 degrees-of-freedom robot).


\section{Related Work}
\subsection{Proxy Collision Checking}

Proxy collision checking is a relatively recent technique that learns a collision classifier and uses it as a proxy for performing geometric collision checks. 
Proxy collision checking methods learn a function $y=\mathrm{sign}(f(x))$ where $y$ is a binary collision label (in-collision or collision-free) and $x$ is a robot configuration. Pan and Manocha explored the use of Support Vector Machines for modeling the boundary between in-collision and collision-free segments of the configuration space \cite{Pan SVM}. They later applied the K-Nearest Neighbors algorithm to the same task with speed improvements of 30$\%$-100$\%$ in motion planning \cite{Pan KNN}. In their work, they also explored parallel collision checks in planning and active learning for dynamic environments. Huh and Lee explored the application of Gaussian Mixture Models to modeling the in-collision and collision-free space \cite{Huh1} \cite{Huh2}. Their methods included transforming the C-space to a circular feature space to account for circular joint values in the C-space and using kinematics to update their mixture models. These methods often make a trade-off between more complex models with higher accuracy or smaller models that can be evaluated faster. 

In contrast to the previously mentioned work, the Fast Perceptron method, Fastron \cite{Fastron Motion Planning}, is a \textit{non-parametric} proxy collision checker which can grow and shrink its model complexity based on the observed complexity of the obstacle boundaries in the C-space. This method was based on the kernel perceptron algorithm and could be trained online, adapt to changing environments, and predict collisions at faster speeds than the previous works. Fastron FK \cite{FastronFK} was a follow-up that defined a Forward Kinematics Kernel Space, which brought the models up to a labeling accuracy of 97$\%$ and could be queried nine times faster compared to state-of-the-art geometric collision checkers. 

\subsection{Configuration Space Decomposition}
A recent work investigated the decomposition of a robot configuration space into subspaces for collision checking. Han et al. proposed a method for decomposing C-space according to degrees of freedom (DOF) of the robot \cite{Dimension Decomp}. They trained a composite classifier of multiple SVMs/KNNs using separate models that incorporated successively larger DOF. While their methods showed improvements in both speed and accuracy, the model's performance was highly dependent on the number of training samples and the number of obstacles in the environment, and they did not report training times for their SVM/KNN-based approach. 


Wong et al. proposed an adaptive method for segmenting C-space for use in GPU-based parallel collision detection \cite{Octree Decomp}. They employed an Octree-based method to subdivide their space depending on the density of obstacles. They did their work in the context of subdividing collision detection for GPUs and not necessarily for proxy collision detection. We compare our method to a generalized version of their Octree-based space decomposition and evaluate the relative performance in our application.

\updated{Quotient spaces are another common configuration space decomposition method which operate on the principle of nested robots with fewer degrees of freedom \cite{Quotient-Space}. While quotient spaces decompose the robot into nested subsets, our method decomposes the environment into disparate subsets, conditioned on the robot. We believe our method could work in conjunction with quotient space motion planners.}

\section{Methods}

\subsection{Configuration Space Decomposition}

Our algorithm starts by decomposing the configuration space into subspaces. To find the decomposed subspaces, uniformly sampled points in the configuration space are first transformed into the Forward Kinematics Kernel Space (FK Space) and then clustered in the FK space.

\begin{figure*}[!t]
    \centering
       \includegraphics[width=0.33\linewidth]{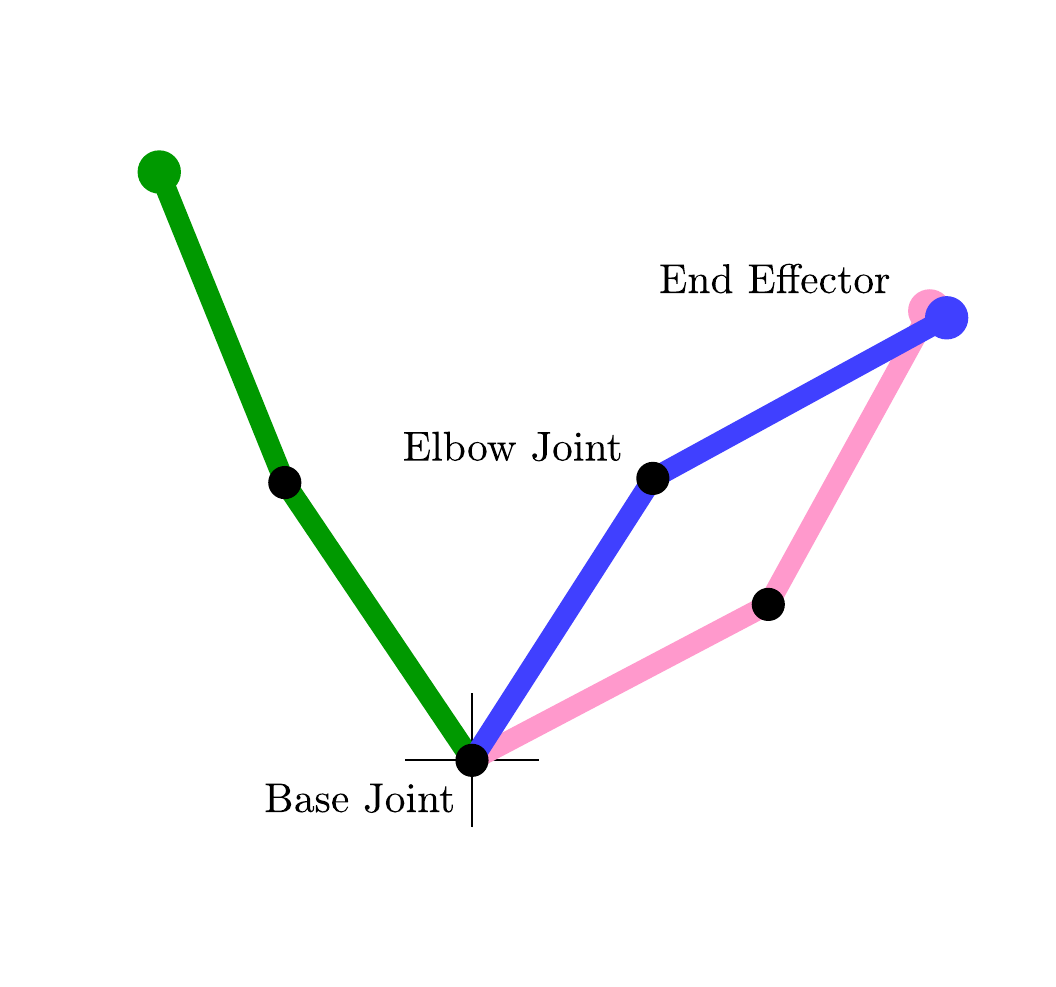}
        \includegraphics[width=0.33\linewidth]{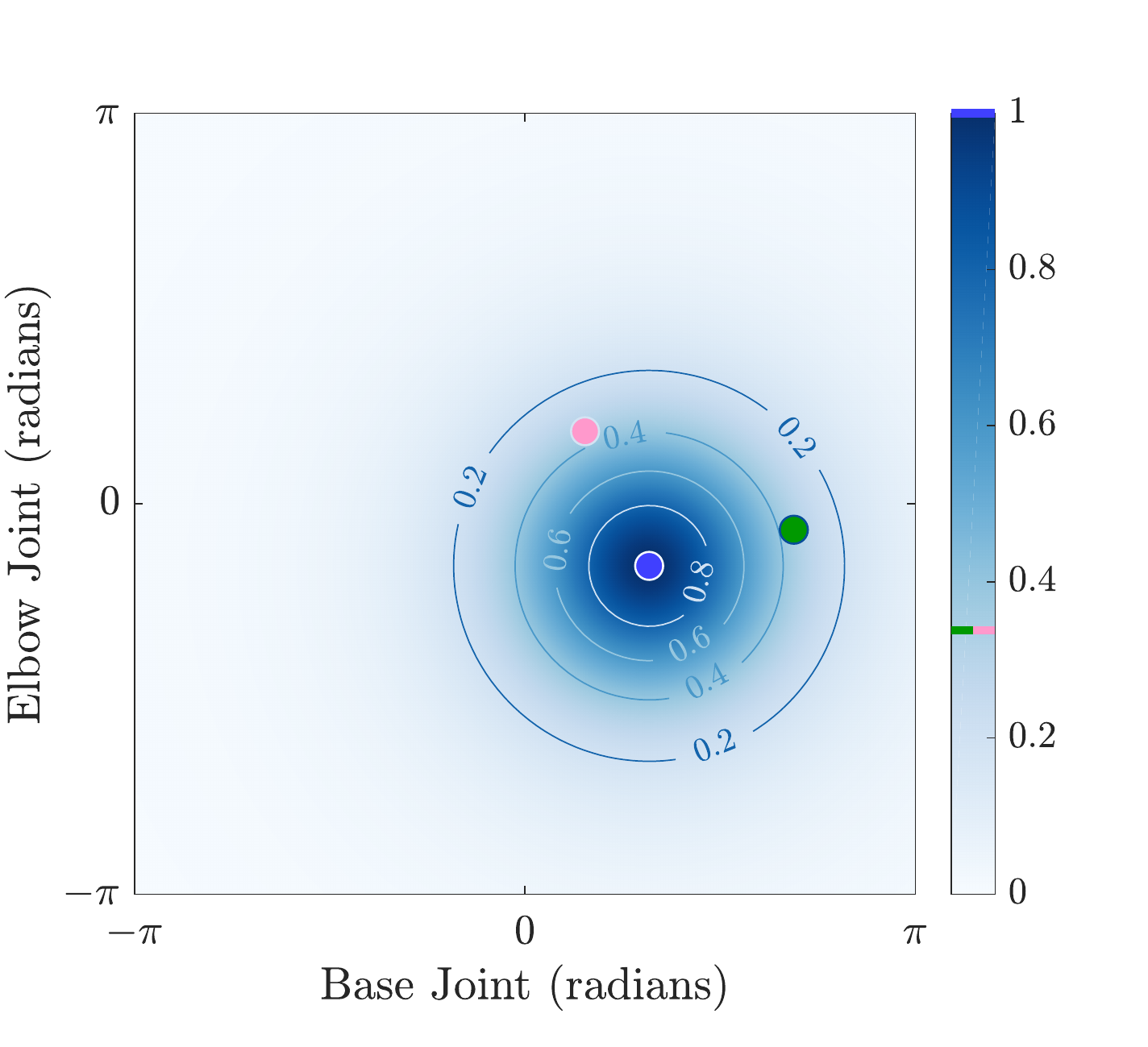}
    \hfill
        \includegraphics[width=0.33\linewidth]{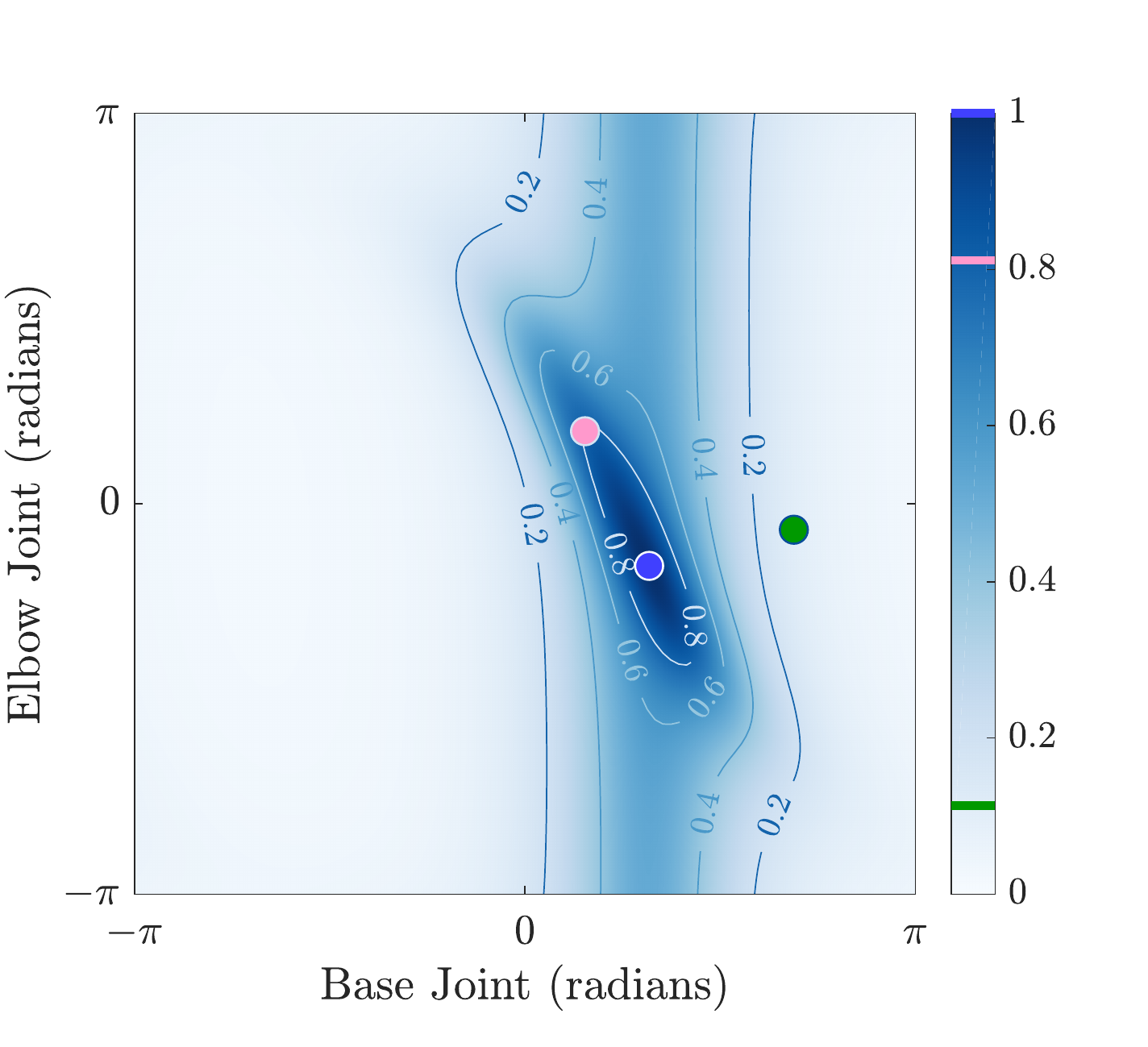}
  \caption{A visualization of how the FK kernel modifies the configuration space map \cite{FastronFK}. Consider a two-link robot arm drawn in several configurations (left). Using a radial basis similarity function (middle), the green configuration is as similar to the blue as the pink is to the blue. Both the green and pink configurations have equal similarity scores. After transforming into the FK Space (right), the blue configuration has a much higher similarity score to pink than to green, which is more realistic.}
  \label{fig:FKspace} 
\end{figure*}

Compared to joint space, distances calculated in the FK space more closely represent distances in Cartesian space.  \updated{Points in an FK-space cluster are more likely to share configuration status, thereby reducing the complexity of a learned model for that cluster.} We construct the FK space by setting $M$ control points on the robot. The position of the $m^{th}$ control point on the robot for a set of joint values $x \in \mathbb{R}^D$ is given by the forward kinematics transform:
\begin{equation}
    \mathrm{fk}_m(x) = \mathrm{pos}({}_w\mathbf{A}_{i_m}\mathbf{T}_m)
\end{equation}
where $i_m$ refers to the nearest joint before the $m^{th}$ control point in the kinematic chain, ${}_w\mathbf{A}_{i_m}$ is a transform from the world frame to that joint frame, and $\mathbf{T}_m$ represents a static transform from joint $i_m$ to the $m^{th}$ control point. We can express our FK space as a concatenation of the workspace locations of all \updated{control points on the robot}:
\begin{equation}
    \mathrm{FK}(x) = 
\begin{bmatrix}
    \mathrm{fk}_1(x) & \mathrm{fk}_2(x) & \ldots & \mathrm{fk}_M(x)
\end{bmatrix}
\end{equation}

Figure \ref{fig:FKspace} (taken from \cite{FastronFK}) highlights the benefits of measuring distance between robot configurations in the FK space. Given the radial basis function as a metric for similarity in the C-space, the green and pink robot configurations are equally similar to the blue configuration. However, in the workspace, the pink and blue configurations are much closer. Measuring similarity in the FK Space represents this property.

To cluster the transformed points we employ K-Means clustering. The K-Means clustering problem is an unsupervised learning problem that seeks to cluster similar points of data in a given space. Given a dataset  $\mathcal{X}$ of $N$ points, $x_i \in \mathbb{R}^D$, we seek $K$ centers $\mathcal{C}, c_i \in \mathbb{R}^D$ to minimize the potential function 
\begin{equation}
    \phi(\mathcal{C}) = \sum_{x\in \mathcal{X}}\min_{c\in\mathcal{C}} \|x-c\|^2,
\end{equation}
the sum of squared distances between points and their closest center. Practically, this is achieved with Lloyd's algorithm shown in algorithm 1.

 Lloyd's algorithm is generally terminated once the decrease in $\phi$ between iterations drops below a certain threshold. Poor initialization of the centers can affect the runtime of the algorithm and the quality of the final clusters at this termination condition. To mitigate this problem, the K-Means++ algorithm \cite{kmeans++} initializes centers more uniformly. Denote the function $Dis(x)$ that gives the shortest distance from $x$ to a previously chosen center. The K-Means++ algorithm picks $c_1$ randomly from $\mathcal{X}$ and then samples every subsequent center $c_i$ with probability $\frac{Dis(x_i)^2}{\sum_{x\in\mathcal{X}} Dis(x)^2}$. Given this better initialization, the K-Means++ algorithm can provide two-fold increases in convergence speed and better final clusters.
 
 \floatstyle{spaceruled}
\restylefloat{algorithm}
\begin{algorithm}[!t]
\vspace{2.5mm}
 \caption{Lloyd's algorithm}
 \begin{algorithmic}[1]
 \renewcommand{\algorithmicrequire}{\textbf{Input:}}
 \renewcommand{\algorithmicensure}{\textbf{Output:}}
 \REQUIRE dataset $\mathcal{X}$
 \ENSURE  centers $\mathcal{C}$, cluster assignments $w$
 \\ \textit{Initialization} : sample initial $K$ centers $\mathcal{C} = [c_1 ... c_K]$
  \WHILE {$\phi_{n-1} > \phi_n$}
    \FOR {$x \in \mathcal{X}$}
    \STATE $w_i = \underset{c\in\mathcal{C}}{\argmin} \|x-c\|^2$
    \ENDFOR
    \FOR{$c \in \mathcal{C}$}
    \STATE $c = \mathrm{mean}(x_i | w_i = c)$
    \ENDFOR
  \ENDWHILE
 \RETURN $\mathcal{C},w$ 
 \end{algorithmic} 
 \end{algorithm}
 
 \begin{figure*}[!t] 
    \centering
  \subfloat[\label{3a}]{%
       \includegraphics[width=0.2\linewidth]{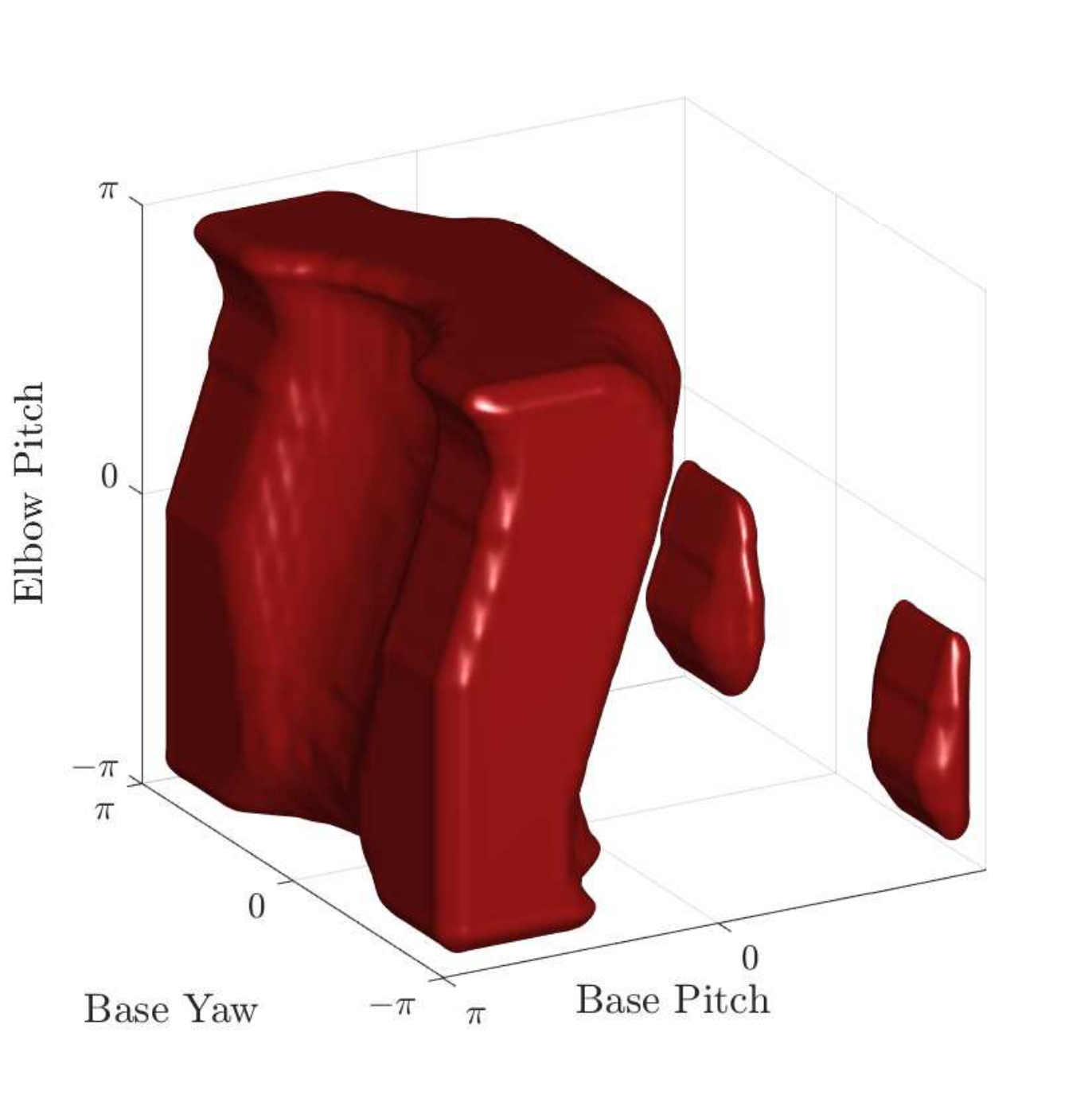}}
  \subfloat[\label{3b}]{%
        \includegraphics[width=0.2\linewidth]{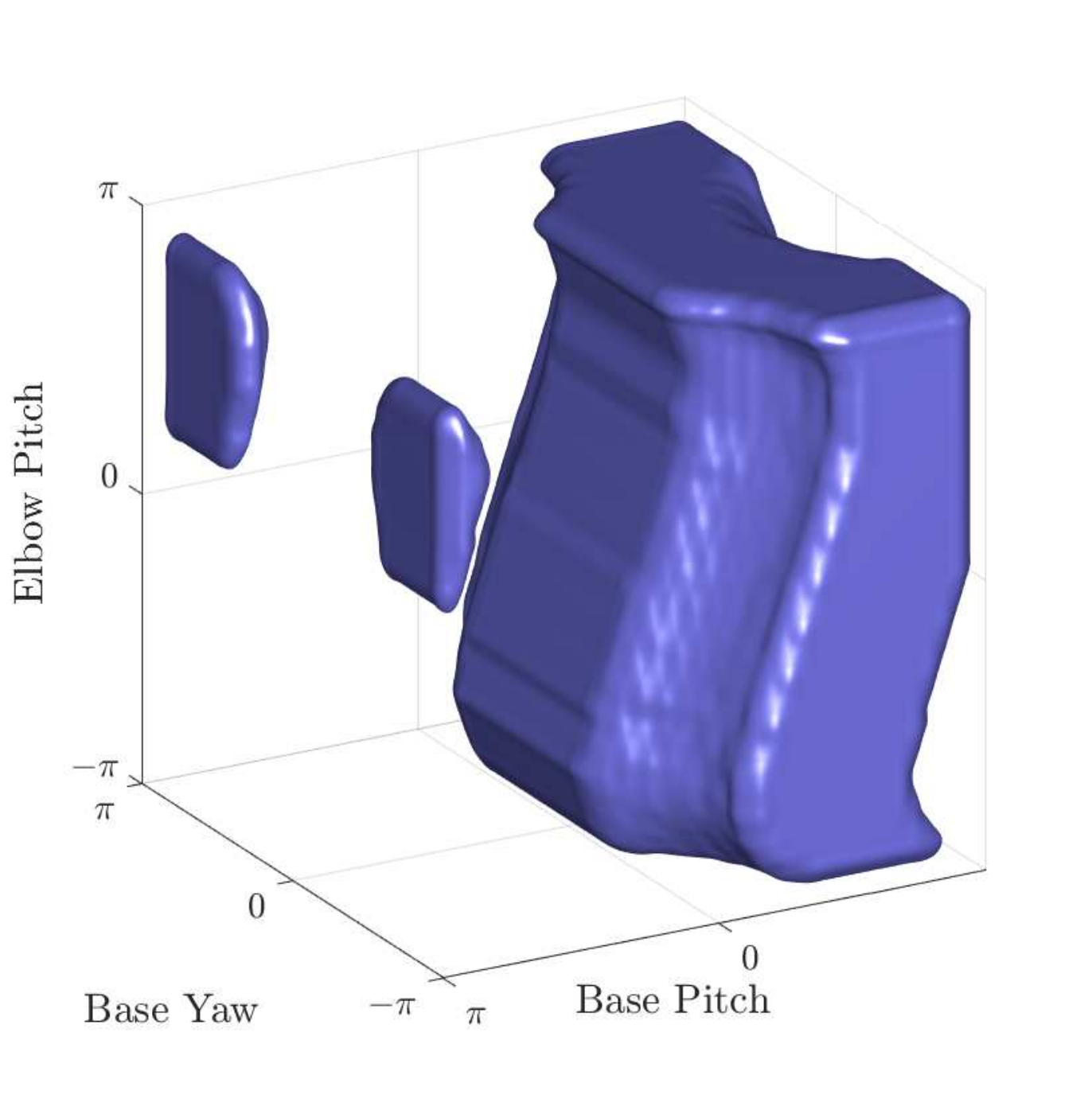}}
  \subfloat[\label{3c}]{%
        \includegraphics[width=0.2\linewidth]{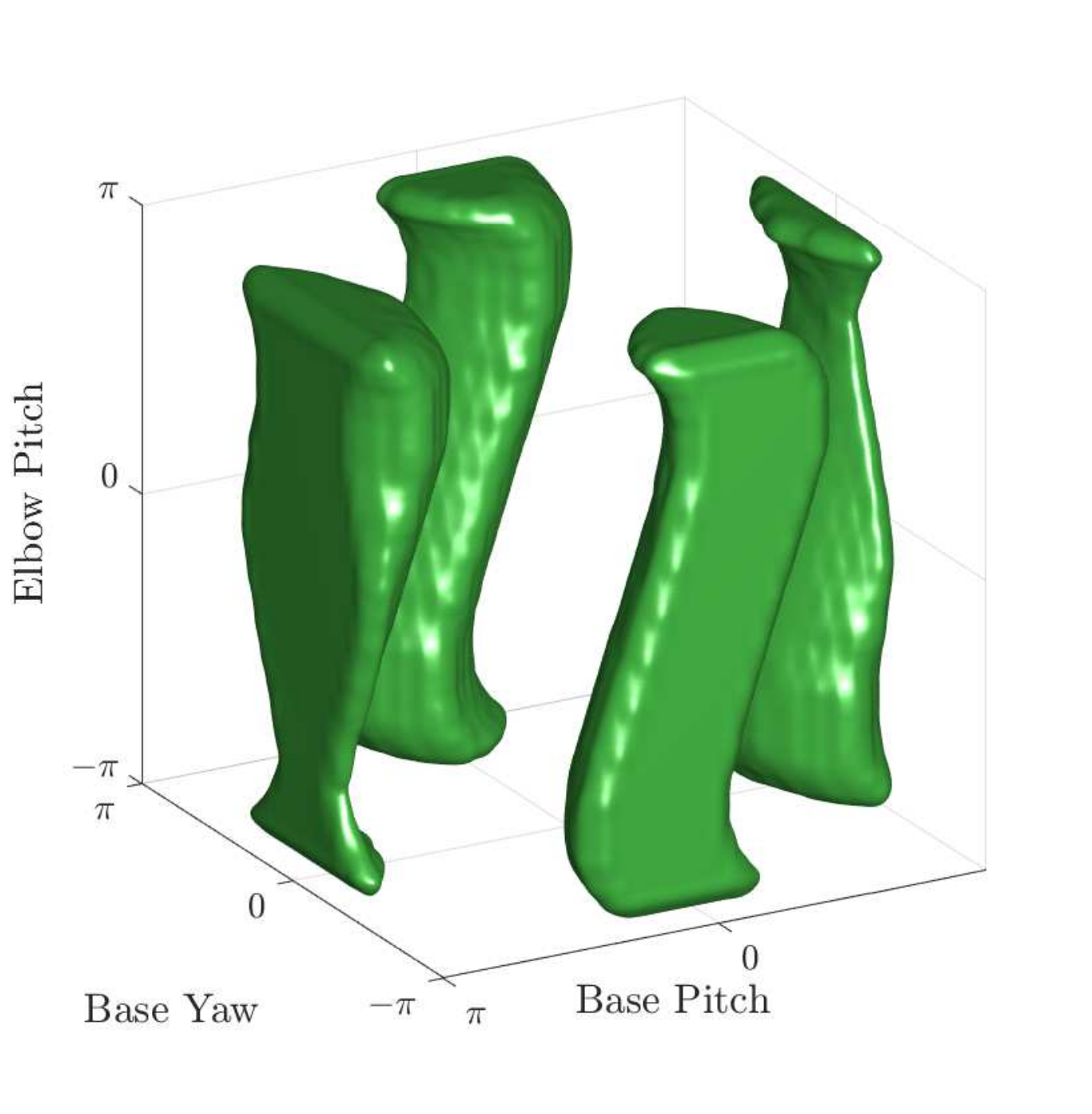}}
  \subfloat[\label{3d}]{%
        \includegraphics[width=0.2\linewidth]{figures/All.pdf}}\\
        \vspace{-4mm}
  \subfloat[\label{3e}]{%
       \includegraphics[width=0.2\linewidth]{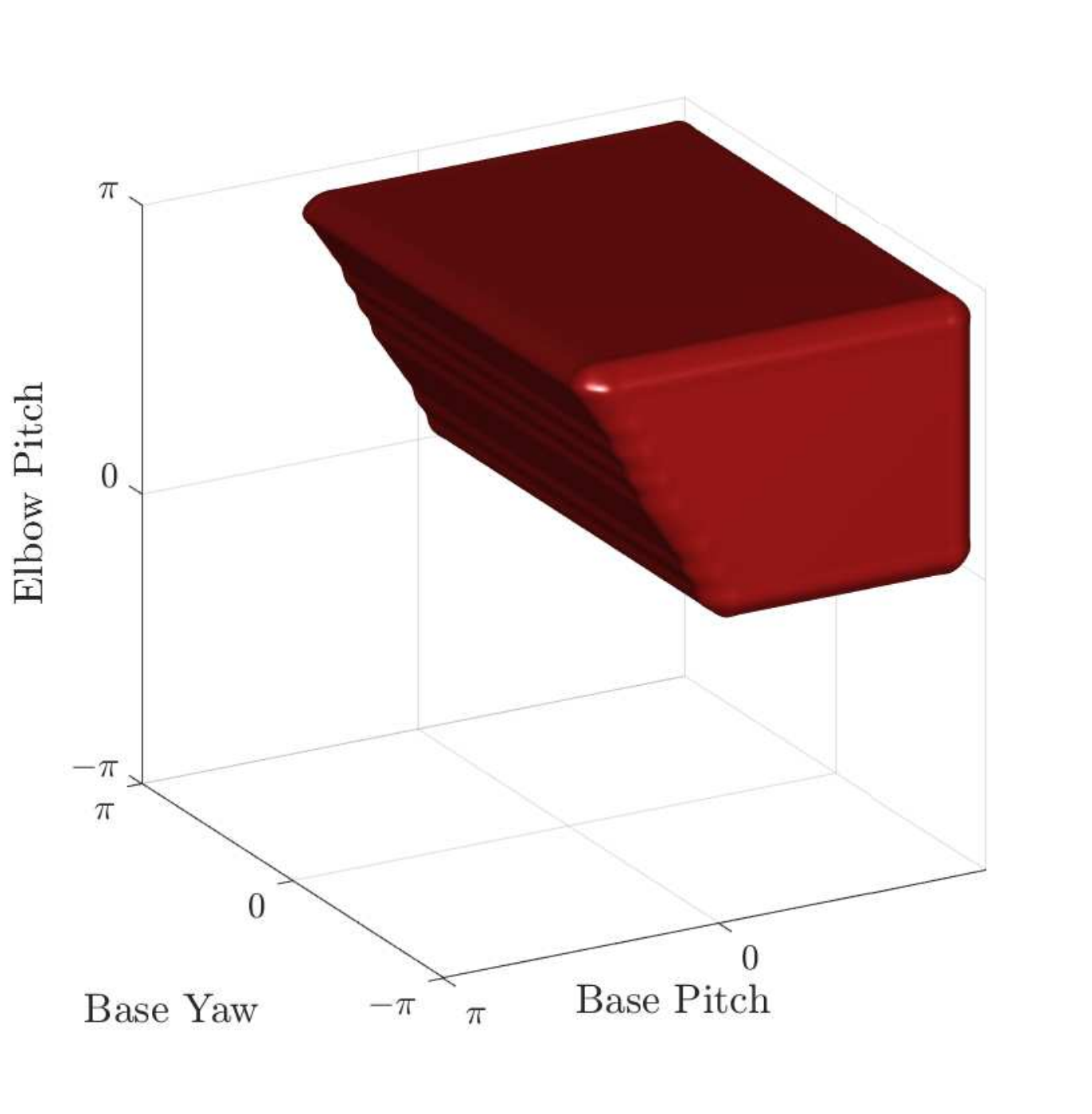}}
  \subfloat[\label{3f}]{%
        \includegraphics[width=0.2\linewidth]{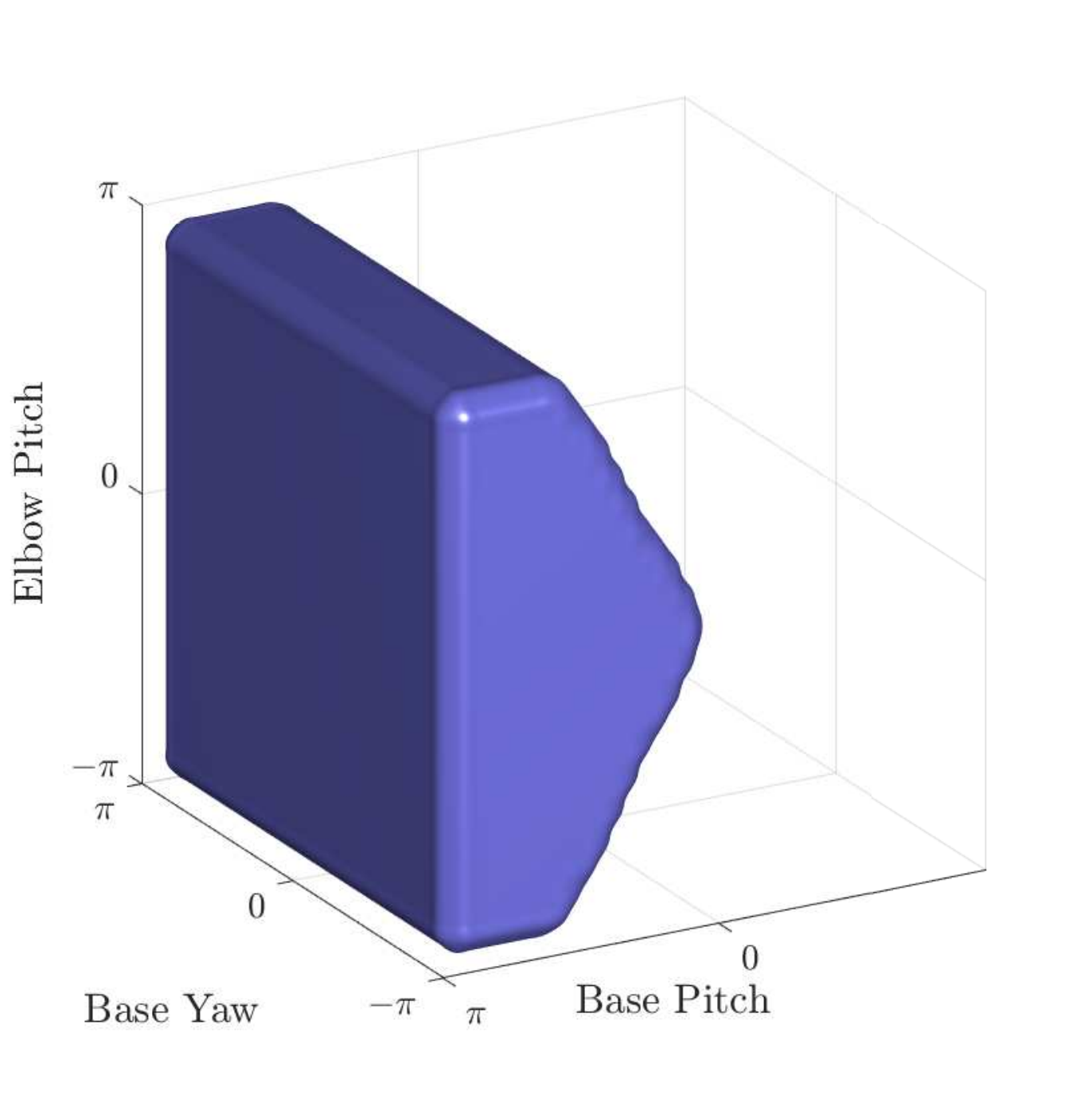}}
  \subfloat[\label{3g}]{%
        \includegraphics[width=0.2\linewidth]{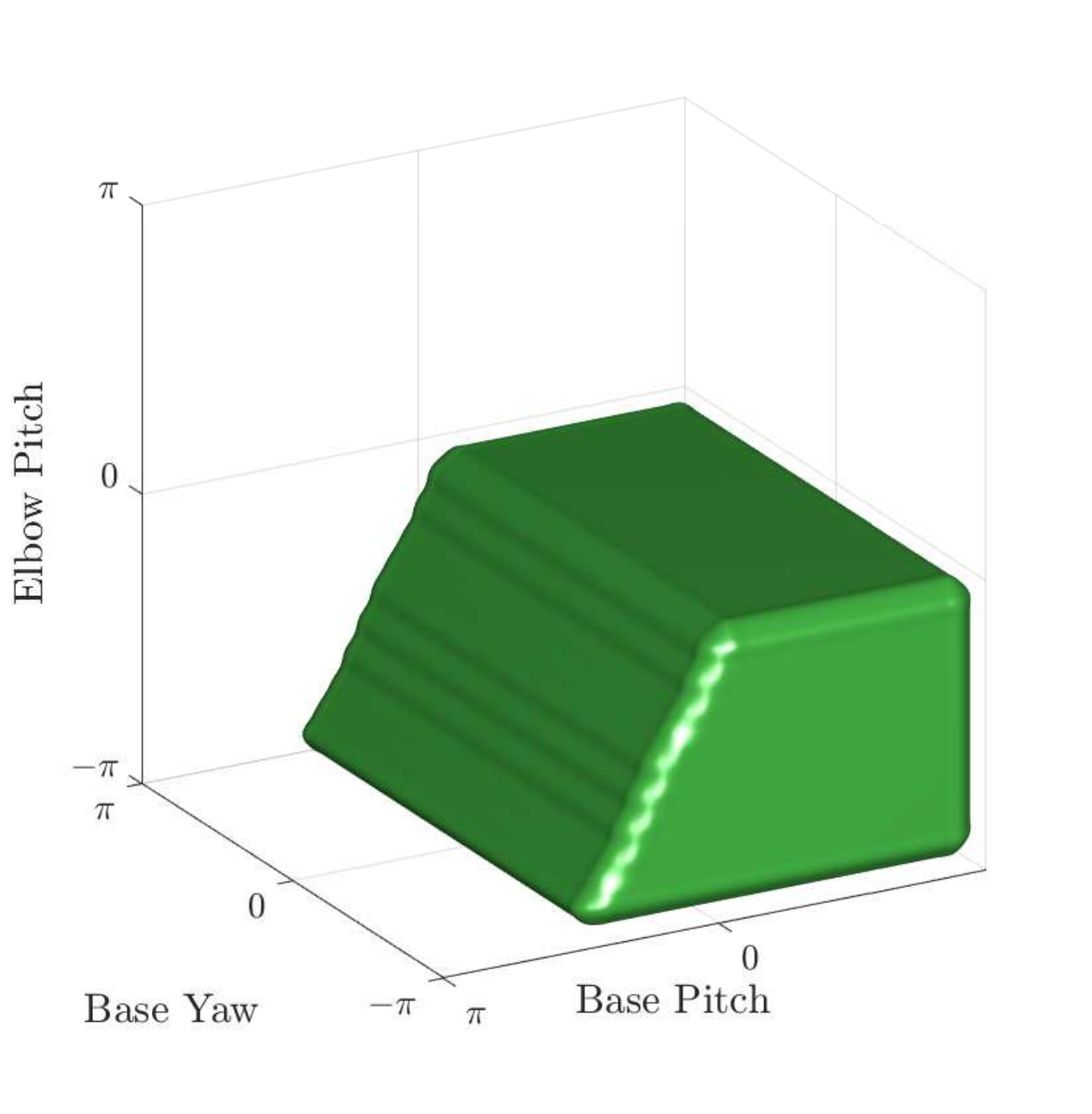}}
  \subfloat[\label{3h}]{%
        \includegraphics[width=0.2\linewidth]{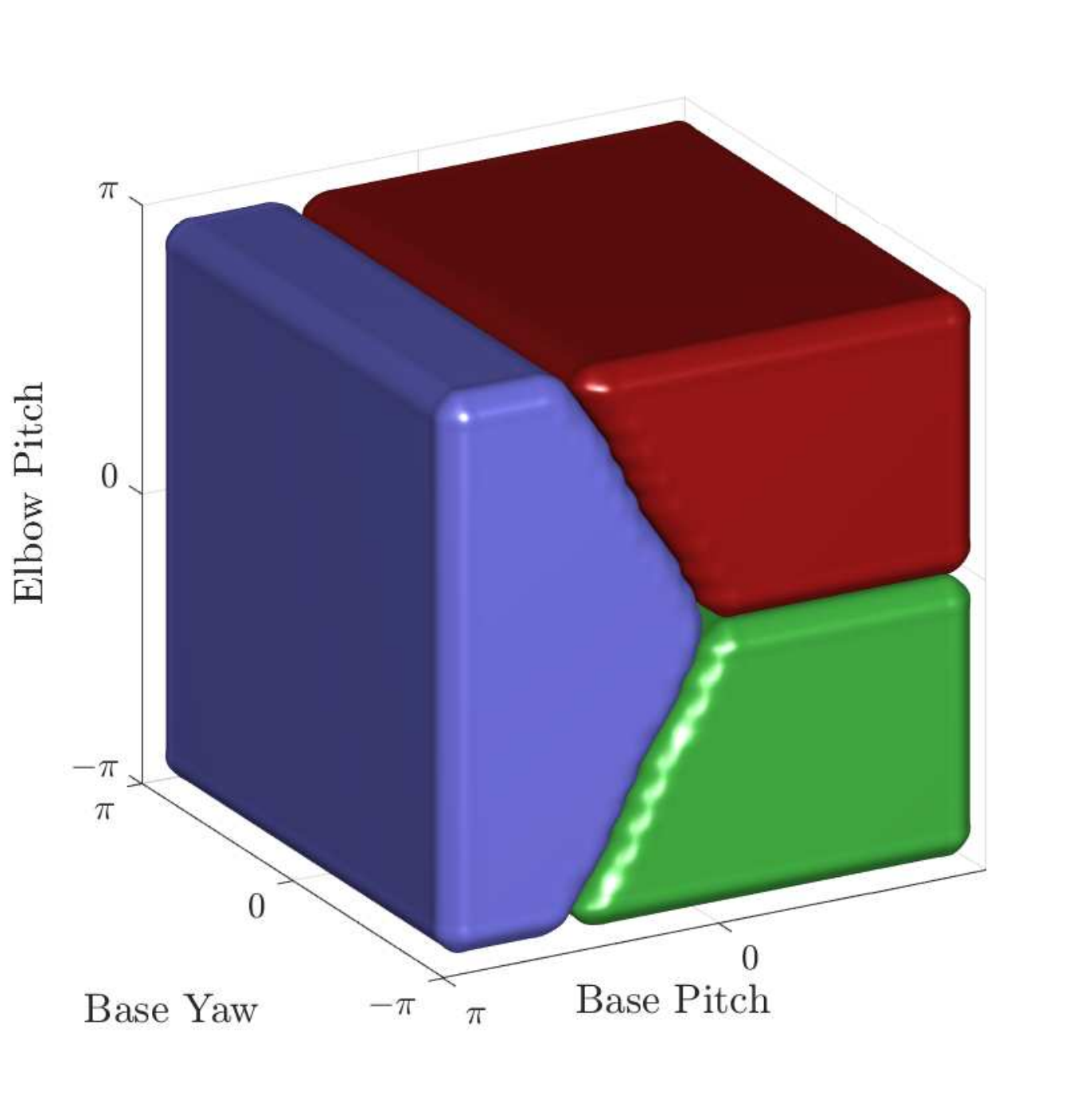}}

  \caption{A comparison of configuration decompositions in different kernel spaces. (a-d) show a decomposition of configuration space by clustering in the FK space for the 3-DOF robot arm using 3 clusters. (a), (b), and (c) each show one of the clusters individually and (d) shows the decomposition of the full space. (e-h) show a naive geometric decomposition of configuration space. Again, (e), (f), and (g) each show one of the clusters individually and (h) shows the decomposition of the full space.}
  \label{fig:decomp} 
   \vspace{-4mm}
\end{figure*}

Given a training data set $\mathcal{X}$, this process of transforming to the FK space and then clustering splits our dataset $\mathcal{X}$ into subsets $\{\mathcal{X}_1, \ldots, \mathcal{X}_K\}$. The centers of these clusters $\{c_1 \ldots c_K\}$ effectively segment the C-space into subspaces which we will call $\{\mathcal{P}_1 \ldots \mathcal{P}_K\}$ where $\mathcal{P}_i$ is the subset of the entire C-space closest to center $c_i$. This decomposition can also be thought of as a Voronoi decomposition based on centers $c_1 \ldots c_K$. For a specific cluster $j$, the set of hyperplanes between $x \in \mathcal{P}_j$ and all other points represents the border of $\mathcal{P}_j$. Figure \ref{fig:decomp} shows these hyperplanes for a $K=3$ decomposition in the C-space of the 3-DOF robotic arm alongside a geometric decomposition done without the FK transform. The nonlinear FK transform produces unique subspaces specifically tailored to the robots kinematics and ultimately results in better performance.

\subsection{The Fastron Algorithm}
 Once the configuration space has been decomposed into subspaces, we employ the Fastron algorithm to train individual configuration space models for each subspace. The Fastron algorithm outlined in detail in \cite{Fastron Motion Planning} trains a binary classifier based on a weighted kernel function (or similarity function) for proxy collision checking. The model takes a set of N training configurations $\mathcal{X} = \{ x_1,...,x_N \}$ where $x_i \in \mathbb{R}^D$ and their corresponding collision labels $y$ where $y_i \in \{-1,+1 \}$ ($+1$ corresponding to a configuration in collision and $-1$ corresponding to a collision-free configuration), and learns a representation of the obstacles in the configuration space. The model is given by:
\begin{equation}
    f(x) = \sum^N_{i=1} K(x_i,x)\alpha_i
\end{equation}
\noindent where $K$ is a positive definite kernel function and $\alpha \in \mathbb{R}^N$ is a learned vector of weights. Proxy collision detection is determined by the $\mathrm{sign}(f(x))$ where $\mathrm{sign}(f(x)) = +1$ indicates the configuration $x$ is predicted to be in collision and $\mathrm{sign}(f(x)) = -1$ is predicted to be collision free. 

The Fastron algorithm employs a rational quadratic kernel evaluated in the FK space dubbed the FK kernel ($K_\mathrm{FK}$). We define the FK kernel as:
\begin{equation}
\begin{aligned}
     K_{\mathrm{FK}}(x,x') &= \frac{1}{M}\sum_{m=1}^M K_{\mathrm{RQ}}(\mathrm{fk}_m(x),~\mathrm{fk}_m(x')), \\
     K_{\mathrm{RQ}}(x,x') &= (1 + \frac{\gamma}{2}\|x-x'\|^2)^{-2},
\end{aligned}
\end{equation}
where $K_\mathrm{RQ}$ is a rational quadratic kernel and $\gamma$ represents the width of the kernel. We also define the kernel matrix $\mathbf{K}$ in an element-wise manner, for each pair of points in $\mathcal{X}:\mathbf{K}_{ij} = K_{\mathrm{FK}}(x_i,x_j)$. This kernel matrix is used later for training. More information on the kernel functions used can be found in \cite{FastronFK}.

The Fastron model is trained using greedy coordinate descent and terminates when the margin $m(x_i) = y_i f(x_i) > 0$ for $\forall x_i \in \mathcal{X}$; when there are no configurations incorrectly classified. When updating weights, the algorithm prioritizes configurations with the largest negative margin. Doing so increases the number of zero weights in the final vector $\alpha$, and generally reduces the number of updates needed for training to converge.


After training, we consider all configurations $x_i$ with nonzero weights $\alpha_i \neq 0$ to be in the set of support points for the model which we denote as $\mathcal{S} \subseteq \mathcal{X}$. As the number of support points directly correlates to the evaluation time for the model, we further prune the support points by removing points from the support set \updated{that are still correctly classified after their removal.}


The Fastron algorithm also contains an active learning algorithm to handle dynamic environments with moving obstacles. The active learning algorithm intelligently \updated{selects points from $\mathcal{X}$} both around existing obstacles to track \updated{their movement}, and then uniformly in \updated{free} space to recognize when new obstacles enter the environment. These points are reclassified by a geometrical collision checker and then used to update \updated{the set of weights} $\alpha$ using the same training algorithm. \updated{The same support point pruning method is applied after active learning updates to keep the model efficient.}


We train an individual Fastron FK model for each subspace in parallel $\{f_1 ... f_K\}$ where $f_i$ is trained on $\mathcal{X}_i$. 
When checking collisions at runtime, we leverage the FK space to compute the closest center $c_i$ to the query configuration $x$ and find the corresponding subspace $\mathcal{P}_i$. We then evaluate the corresponding Fastron model $y = f_i(x)$ to determine the collision status of the queried point. The time to localize the new configuration into a subspace adds overhead and could outweigh potential computational gains from using smaller models given by the decomposition strategy. However, the original Fastron FK algorithm already computes the FK transform as part of its queries, making that operation computationally free to us. As a result, the only computational impact incurred by our localization step is the distance computation between the query point and the cluster centers. 

\updated{Our final training algorithm transforms the dataset to the FK space, clusters the transformed dataset and segments each cluster into its own subset, and finally trains a Fastron model on each subset. This process is detailed in Algorithm 2. Evaluating a single collision check requires localizing a point to its appropriate subspace by computing the closest cluster center in the FK space, and then querying the relevant model. To leverage the efficient matrix computation found in the Fastron algorithm, when checking batches of points, we first localize all points into subsets and then perform batch model queries on each of these subsets. This batched model evaluation is shown in Algorithm 3.}

\begin{algorithm}[!t]
 \caption{D-Fastron Training}
 \begin{algorithmic}[1]
 \renewcommand{\algorithmicrequire}{\textbf{Input:}}
 \renewcommand{\algorithmicensure}{\textbf{Output:}}
 \REQUIRE dataset $\mathcal{X} \in \mathbb{R}^{N \times D}$ and labels $y \in \{-1, +1\}^{N}$
 \ENSURE  set of models $\{f_1 ... f_K\}$
 \STATE $\mathcal{X}_\text{fk}\in \mathbb{R}^{N \times3M} = \mathrm{FK}(\mathcal{X})$
 \STATE $\mathcal{C} \in \mathbb{R}^{K\times3M},w\in{\mathbb{Z}_K}^{N} = \text{K-Means++}(\mathcal{X}_\text{fk})$
 \STATE split $\mathcal{X}_\text{fk} \rightarrow \{\mathcal{X}_1 ... \mathcal{X}_K\}$ using cluster assignments $w$ 
 \STATE split $y \rightarrow \{y_1 ... y_K\}$ using cluster assignments $w$
 \FOR{each $\mathcal{X}_i$}
    \item train $f_i(x)$ using $\mathcal{X}_i$ and $y_i$ 
 \ENDFOR
 \RETURN $\{f_i ... f_K\}, \mathcal{C}$ 
 \end{algorithmic} 
 \end{algorithm}

 \begin{algorithm}[!t]
 \caption{D-Fastron Evaluation}
 \begin{algorithmic}[1]
 \renewcommand{\algorithmicrequire}{\textbf{Input:}}
 \renewcommand{\algorithmicensure}{\textbf{Output:}}
 \REQUIRE dataset $\mathcal{X} \in \mathbb{R}^{N \times D}$, models $\{f_1 ... f_K\}$, and cluster centers $\mathcal{C} \in \mathbb{R}^{K\times3M}$
 \ENSURE  prediction $y' \in \{-1, +1\}^{N}$
 \\ \textit{Initialization} : set of empty subsets, \{$\mathcal{X}_1$ ... $\mathcal{X}_K$\} \\
 \STATE $\mathcal{X}_\text{fk} \in \mathbb{R}^{N\times3M} = \text{FK}(\mathcal{X})$
 \FOR{$x \in \mathcal{X}_\text{fk}$}
    \STATE $i = \underset{i}{\argmin} \|x-C_i\|^2$
    \STATE $ \mathcal{X}_i \leftarrow \mathcal{X}_i \cup \{x\}$
 \ENDFOR
 \FOR{\updated{$i \in \{1,\dots,K\}$}}
    \STATE $y_i' = f_i(\mathcal{X}_i)$
 \ENDFOR
 \updated{\STATE $y' = \bigcup\limits_{i=1}^{K} y_i'$}
 \RETURN $y'$ 
 \end{algorithmic} 
 \end{algorithm}
 
 In practice, we use parallel computation for both the model training and the active learning algorithm to minimize increases in training time. While different models in different subspaces may train and update faster, we rely on the operating system to schedule training on threads in the most efficient manner. Furthermore, during model evaluation, we could check batches of points at the same time via parallelization across CPUs or GPUs. We leave such optimizations for future work. 
 
 \section{Results}
 We performed our experiments in simulation using the 7-DOF right arm of the Baxter robot from Rethink Robotics. To validate our K-Means-based D-Fastron model, we compare it to the vanilla Fastron FK model and another decomposed Fastron FK model that decomposes space in a method inspired by \cite{Dimension Decomp} and \cite{Octree Decomp} involving bisection. This third method bisects each degree of freedom of the robot to create a tree of subspaces. Starting at the shoulder joint and moving down the Baxter's arm, we bisect $n$ joints giving the tree $2^n$ subspaces. This method decomposes the space into anywhere from 2 to 128 subspaces. For practicality, we test decompositions ranging from 4 to 32 subspaces for both the K-Means-based decomposition and the bisection decomposition. We place the control points for the FK transform at joint frame origins (according to Denavit-Hartenberg conventions), except where placements would be redundant. Functionally, this results in four control points, one at the distal end of each link in the Baxter's arm. More information about Denavit-Hartenberg parameters for the FK transform can be found in \cite{FastronFK} and more information about the Baxter's specific kinematics can be found in \cite{BaxterFK}. Initially, we trained each model with a dataset of 10,000 configurations labeled with collision statuses and three randomly placed cuboid obstacles in the arm's workspace. We trained each of the models with the same sampled dataset and the same obstacle placements and size. We also use the same Fastron model parameters for every method. The Fast Collision Library (FCL) \cite{FCL} that comes in the ROS Open Motion Planning Library provided collision labels for the data.

 \subsection{Training and Model Complexity}
\begin{figure}[tb] 
    \centering
       
       \includegraphics[width=.90\linewidth,trim={1cm 0 2cm 1cm},clip]{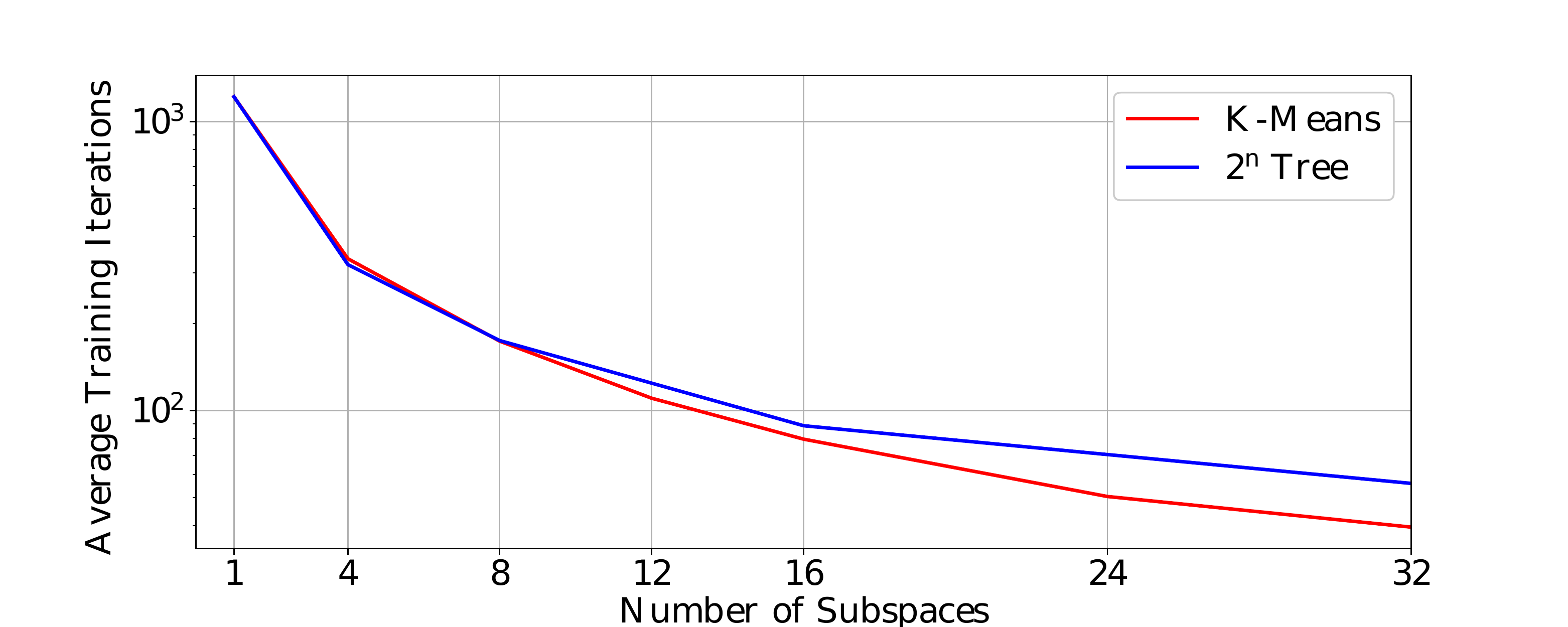}\\
       (a) Average Number of Training Iterations\\
        \includegraphics[width=.90\linewidth,trim={1cm 0 2cm 1cm},clip]{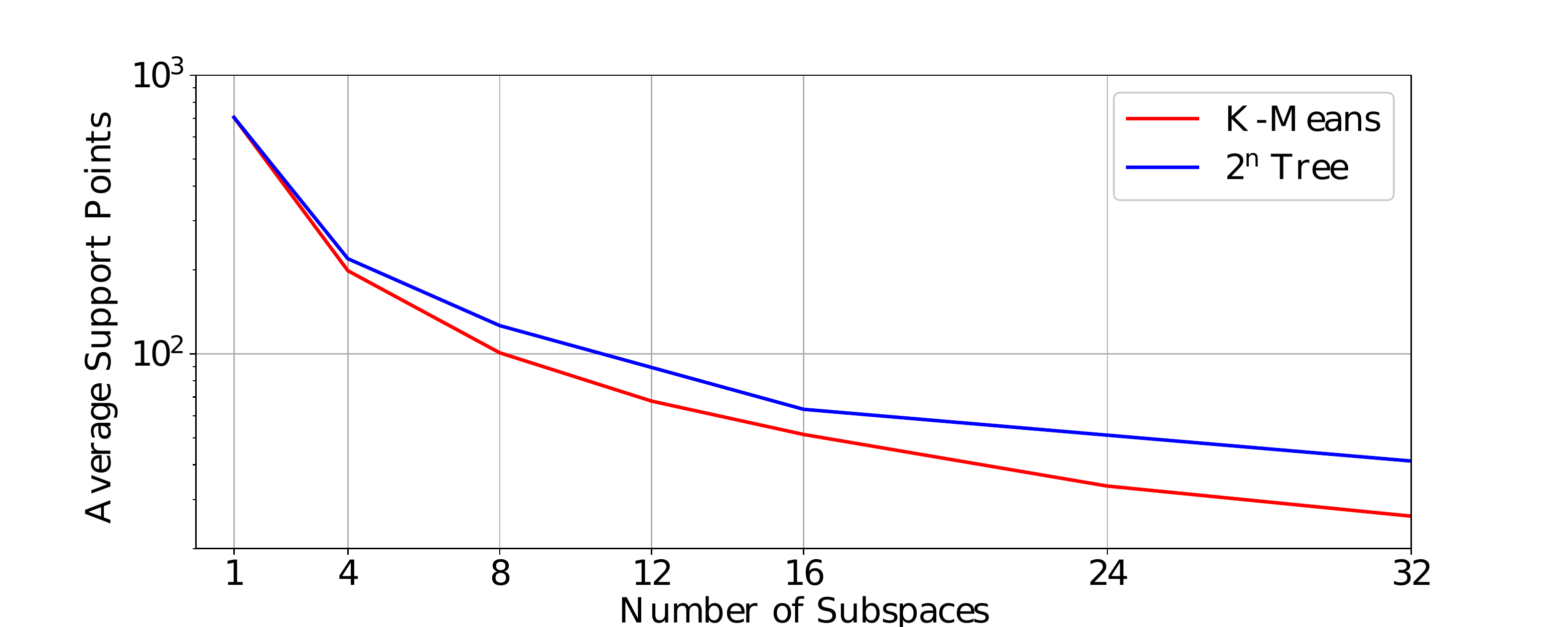}\\
         (b) Average Number of Support Points Per Decomposition
  \caption{Plots of average training iterations (a), and average number of support points (b), with respect to the number of decompositions of the configuration space. Note that 1 subspace is equivalent to the base Fastron FK model. }
  \label{fig:Training} 
\end{figure}

 We measured the average number of training iterations of each of the models to quantify changes in training time for our algorithm (see Fig. \ref{fig:Training}). Note in the plots that the leftmost data point corresponding to one subspace is the original Fastron FK algorithm with no space decomposition. For fewer models, the K-Means and bisection decomposition methods were comparable. As the number of models increases, the K-Means method outperformed the bisection method and trained 30$\%$ faster with a 32 subspace decomposition. Both models initially show training speedups proportional to the order of the space decomposition but with overall diminishing returns for highly decomposed spaces.

  \begin{figure}[!b] 
    \centering
    \includegraphics[width=.8\linewidth, trim={1cm 3cm 1cm 7.5cm}, clip]{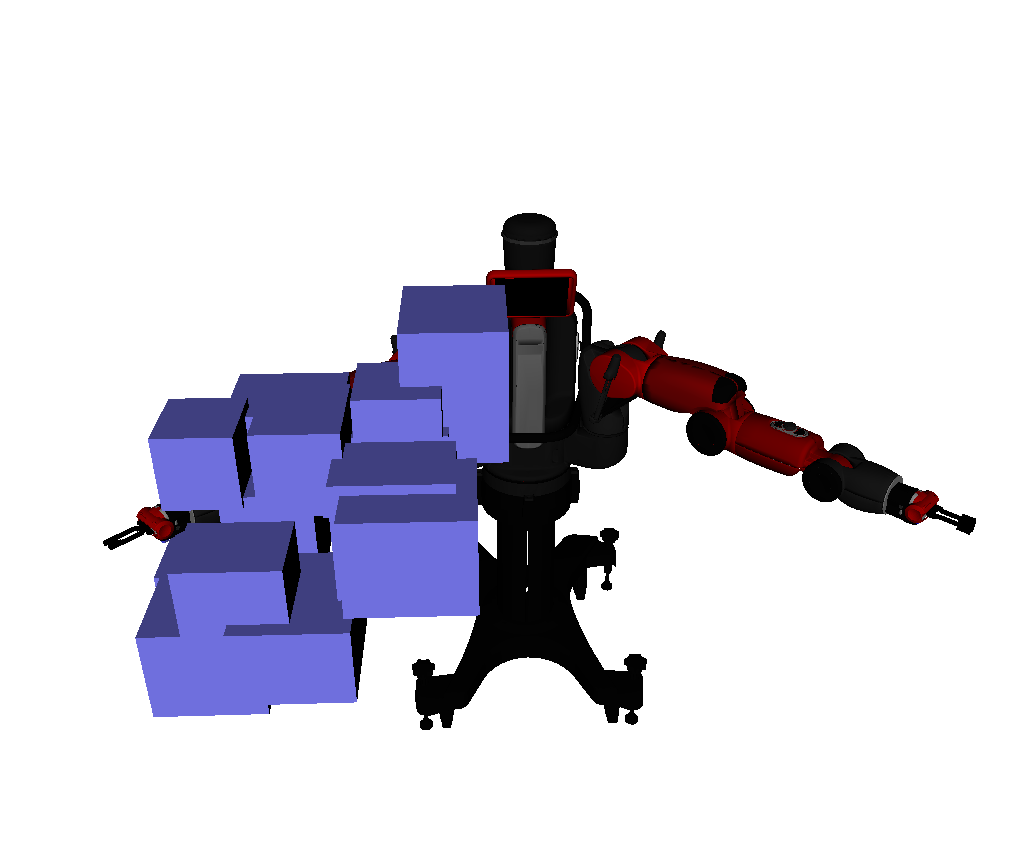}
   \caption{The Baxter robot used in motion planning evaluations with 15 obstacles in its workspace. The obstacles comprise 42\% of the right arm's workspace.}
  \label{fig:Baxter} 

\end{figure}
 
 We also examine the average number of support points as a heuristic for model complexity, and ultimately model evaluation time. The K-Means decomposition method outperforms the bisection method producing models with 37$\%$ fewer support points with a 32 subspace decomposition. This result indicates the K-Means method creates a fundamentally better decomposition of the configuration space, likely due to its consideration of robot kinematics. At a 32 subspace decomposition, the K-Means method trains models that have on average 27 times fewer support points than the base Fastron FK.
 
 \subsection{Collision Checking Timings}
\begin{figure}
    \centering
       \includegraphics[width=.90\linewidth,trim={1cm 0 2cm 1cm},clip]{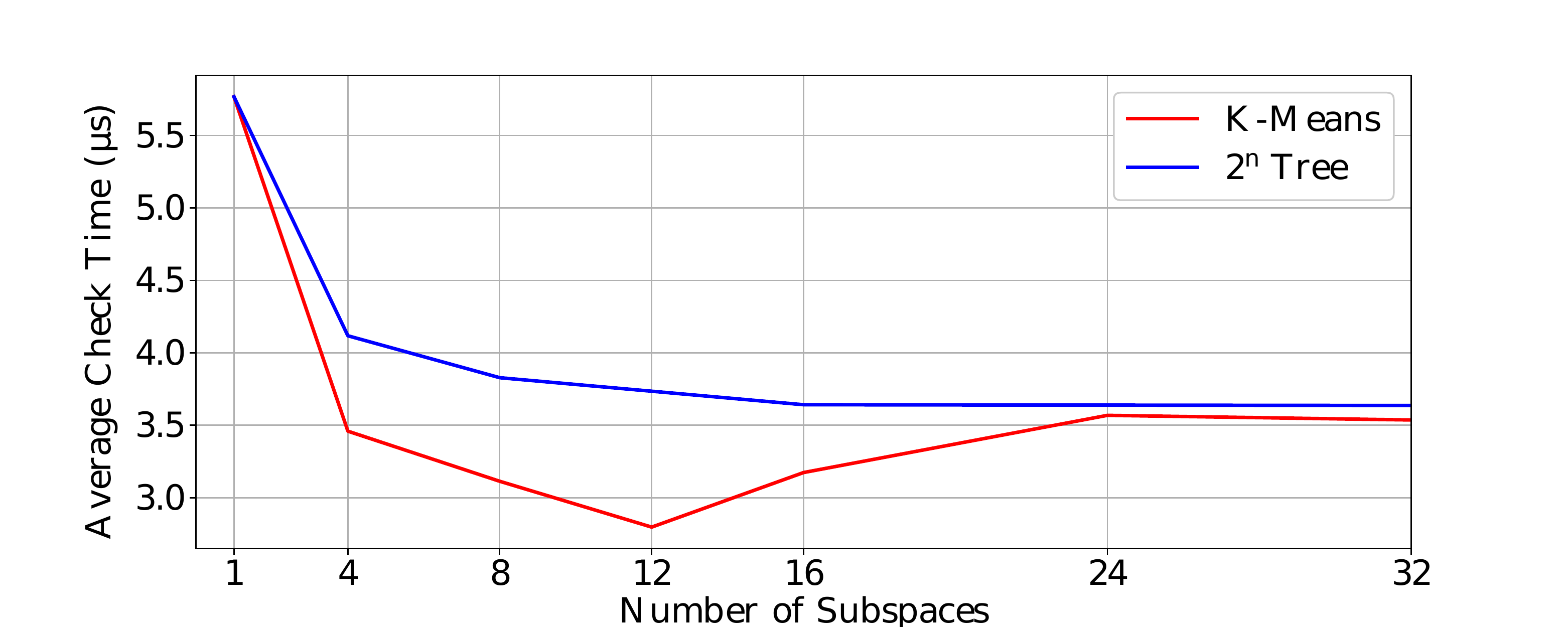}
     \caption{Plot of collision checking times with respect to the number of subspaces the configuration space is decomposed into. 1 subspace is equivalent to the base Fastron FK model.}
  \label{fig:Testing} 
  \vspace{-3mm}
\end{figure}

While examining model complexity showed significant improvement in the decomposed Fastron methods over Fastron FK, it is important to note these improvements will not translate linearly into collision checking speedups. The FK transform can be an expensive operation and adds overhead to the Fastron FK model. In addition, both decomposition methods add time to localize a point into its corresponding subspace on top of this overhead.  Figure \ref{fig:Testing} shows the collision checking times for the base Fastron FK and the two space decomposition methods. We obtained these values as an average of 30 successive runs of 100,000 collision checks each to normalize for fluctuations in compute time. The base Fastron FK model averaged 5.76 microseconds per collision check. The bisection decomposition method appeared to hit an asymptote of 3.64 microseconds per collision check. Testing further decompositions did not yield any further improvement. The K-Means decomposition method peaked at 2.79 microseconds per collision check with 12 models, a full 2-fold faster than the base Fastron FK model. As the bisection method employs a binary tree, the complexity to localize points scales as $O(\log K)$ where $K$ is the number of subspaces. The K-Means-based method has to calculate distances between the query point and every cluster center, and its localization time only scales as $O(K)$. Despite the K-Means based method's longer localization time, its fundamentally better space decomposition gives the method an overall advantage. Going forward, we will be only comparing the K-Means-based decomposed Fastron with 12 subspaces (to be referred to as D-Fastron) as we have shown it to be the superior decomposed Fastron method.

\begin{table}
\vspace{1.5mm}
\caption{Average collision checking performance of D-Fastron, Fastron FK and GJK across varying numbers of obstacles in the environment. Lower is better for query time, support points, and model update time.}
\label{table:Check Timings}

\footnotesize
\setlength{\tabcolsep}{6pt}
\centering
\ra{1.3}
\resizebox{\linewidth}{!}{
\begin{tabularx}{\linewidth}{@{}Xcccc@{}}
\toprule
Method    & $\#$ Obs & Query Time & Support Pts & Update Time \\
\midrule
   & 3 & 2.71\ $\mu$s & 50 & 13.10\ ms \\
 D-Fastron           & 4 & 2.75\ $\mu$s & 69 & 18.23\ ms \\
            & 5 & 3.63\ $\mu$s &  88 & 21.18\ ms \\
\midrule
  & 3 & 5.41\ $\mu$s& 567& 8.11\ ms \\
Fastron FK            & 4 & 6.82\ $\mu${s} & 644 & 19.26\ {ms} \\
            & 5 & 7.99\ $\mu$s& 1021& 29.84\ ms \\
\midrule
         & 3 & 43.88\ $\mu$s & --- & --- \\
GJK            & 4 & 46.33\ $\mu$s & --- & --- \\
            & 5 & 51.12\ $\mu$s & --- & --- \\

\bottomrule
\end{tabularx}
}
\end{table}

Table \ref{table:Check Timings} shows the query times, model complexity as represented by the average number of support points, and update timings for D-Fastron, a single Fastron FK model, and GJK, a common approximate convex geometric collision detection algorithm \cite{GJK}. Each algorithm was tested across varying numbers of cubiod obstacles with sides randomly sized between 0.2-0.4 meters and distributed across the workspace. Note that the query timings here are inclusive of all sorting times and represent the average time for a full collision check. We also tested update times for active learning. Obstacles were given a random direction and moved for 30 time steps in that direction. Active learning updates were performed after each time step and the update times were averaged. Despite training more models, D-Fastron's update time is comparable to Fastron FK due to parallel training. The fast model update times ensure D-Fastron can be run online in dynamic environments at reasonable refresh rates.

\begin{figure}
    \centering
       \includegraphics[width=.90\linewidth,trim={1cm 0 2cm 1cm},clip]{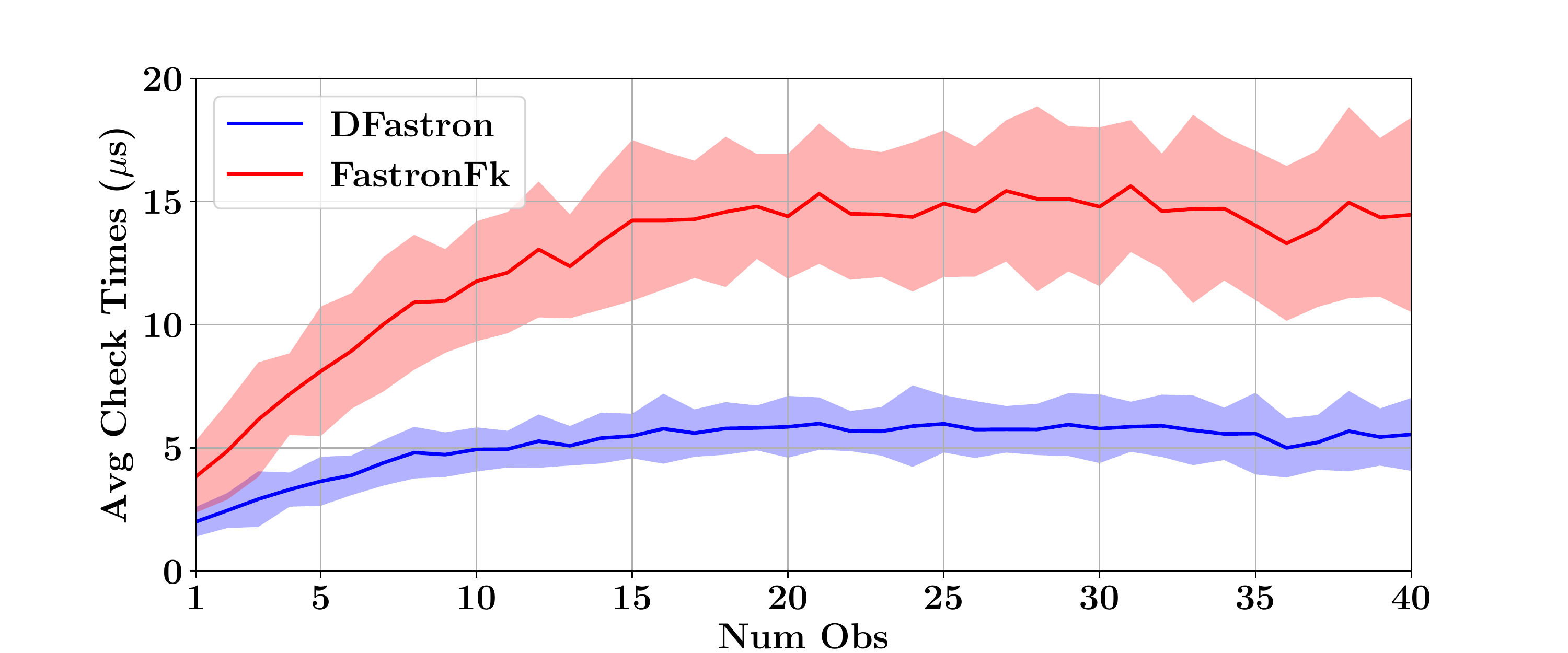}
     \caption{Plot of average collision checking times with respect to the number of obstacles in the configuration space for D-Fastron and Fastron FK. The shaded regions show one standard deviation in the collected data.}
  \label{fig:40ObsColCheck} 
  \vspace{-3mm}
\end{figure}

We tested D-Fastron, and Fastron FK in environments with 1 to 40 obstacles to see how their collision checking performance scaled. For each number of obstacles, we generated 50 random obstacle placements and tested each collision checking method on 100,000 points. These results can be seen in Figure \ref{fig:40ObsColCheck} where the solid lines show the average collision checking time and the shaded portions show one standard deviation in check times. D-Fastron scales significantly better in query time with respect to the number of obstacles in the workspace compared to a single Fastron FK model. Overall, D-Fastron demonstrates a collision checking speed ranging from 1.9-2.7$\times$ faster \updated{than} Fastron FK, averaging a 2.46$\times$ improvement. D-Fastron also demonstrates a significant average 29$\times$ speedup over GJK \cite{GJK}, one of the fastest geometric collision checkers.




\subsection{Model Correctness}
Model correctness was measured through overall classification accuracy and true positive rate (TPR), both of which are important for collision checking and motion planning. The classification accuracy and TPR are defined as:
\begin{equation}
\begin{aligned}
     \text{Accuracy} &= \frac{TP + TN}{TP + FN + TN + FP}\\
     \text{TPR} &= \frac{TP}{TP+FN}
\end{aligned}
\end{equation}
where $TP$ and $TN$ are the number of samples correctly predicted to be in collision and collision free respectively, and $FP$ and $FN$ are the number of samples falsely predicted to be in collision and collision free respectively.
A high TPR is crucial to minimize the amount of path repair needed when planning \cite{Fastron Motion Planning}. When the dataset has an imbalance between in-collision and collision free points, examining the TPR helps give a better insight into the model's correctness.

\begin{figure}[tb] 
    \centering
       
       \includegraphics[width=.90\linewidth,trim={1cm 0 2cm 1cm},clip]{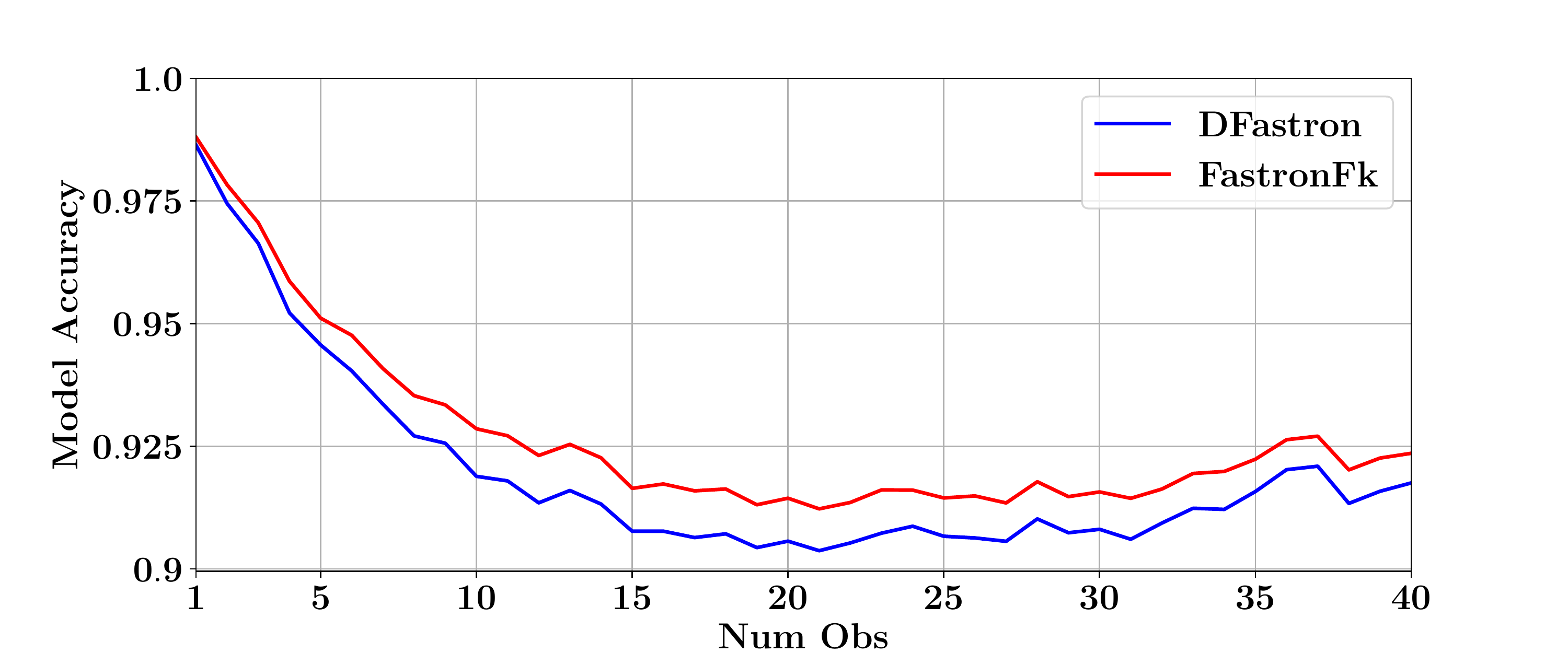}\\
       (a) Model Accuracy\\
        \includegraphics[width=.90\linewidth,trim={1cm 0 2cm 1cm},clip]{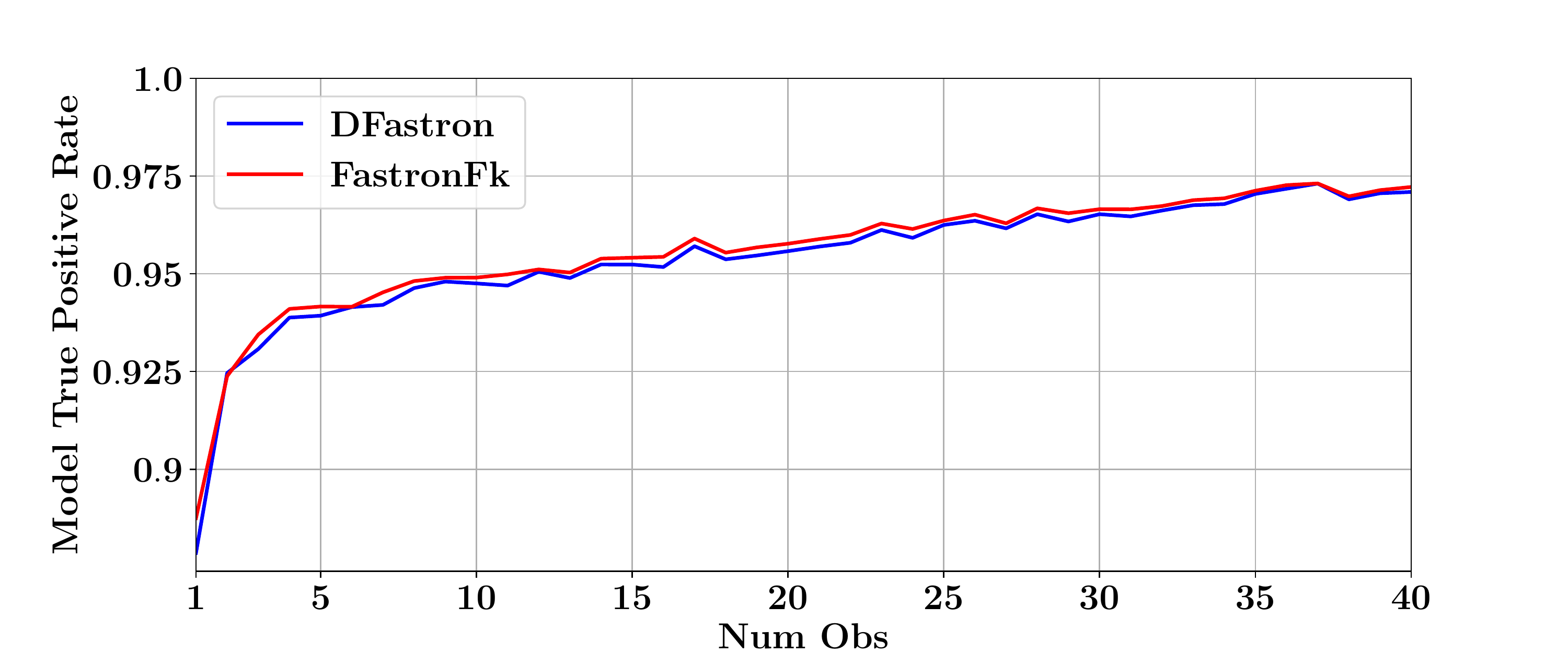}\\
         (b) Model True Positive Rate
  \caption{Plots of average model accuracy (a), and average model true positive rate (b), with respect to the number of obstacles in the workspace for D-Fastron and Fastron FK. }
  \label{fig:ModelCorrectness} 
\end{figure}

The model correctness is shown in Figure \ref{fig:ModelCorrectness}. As the number of obstacles in the workspace increase, the true positive rate naturally increases, and the accuracy (and true negative rate) decrease slightly to compensate. The D-Fastron model correctness is comparable to Fastron FK never deviating by more than a percent \updated{despite each subspace model being trained with much less data than a global model.}

\subsection{Motion Planning}
\begin{figure*}[tb] 
    \centering
       
       \includegraphics[width=.3\linewidth,trim={0cm 0cm 0cm 1cm},clip]{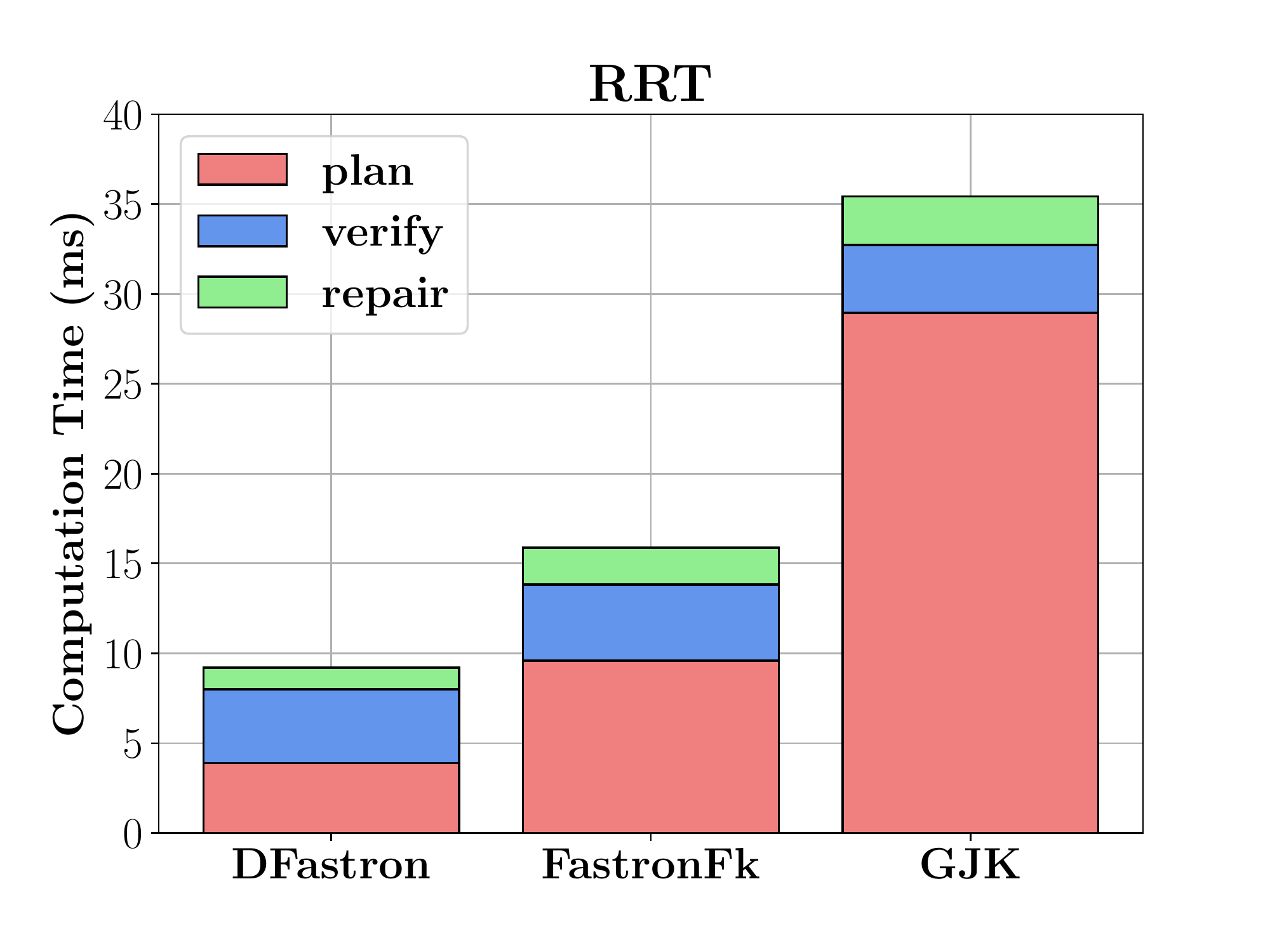}
        \includegraphics[width=.3\linewidth,trim={0cm 0cm 0cm 1cm},clip]{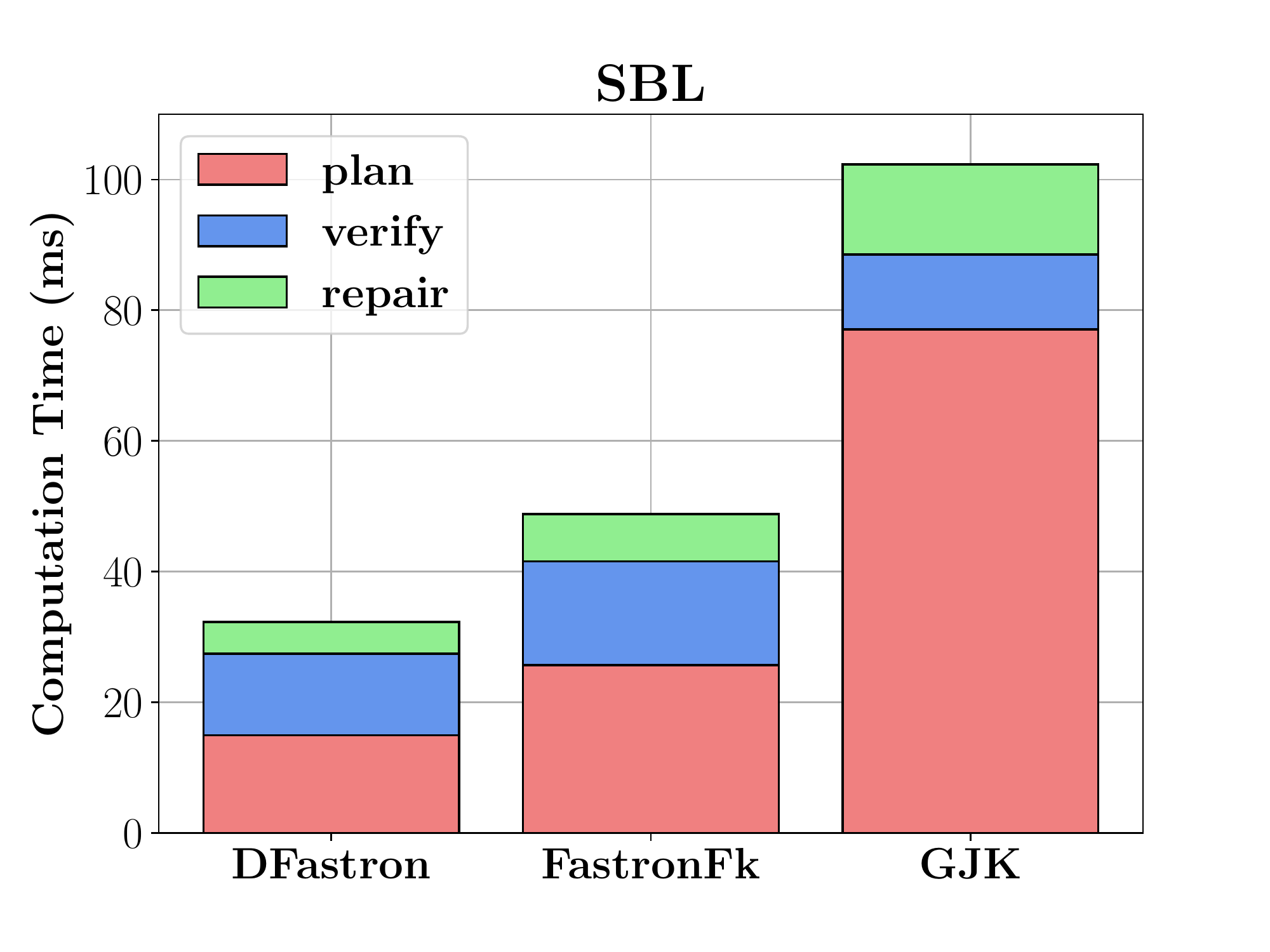}
       \includegraphics[width=.3\linewidth,trim={0cm 0cm 0cm 1cm},clip]{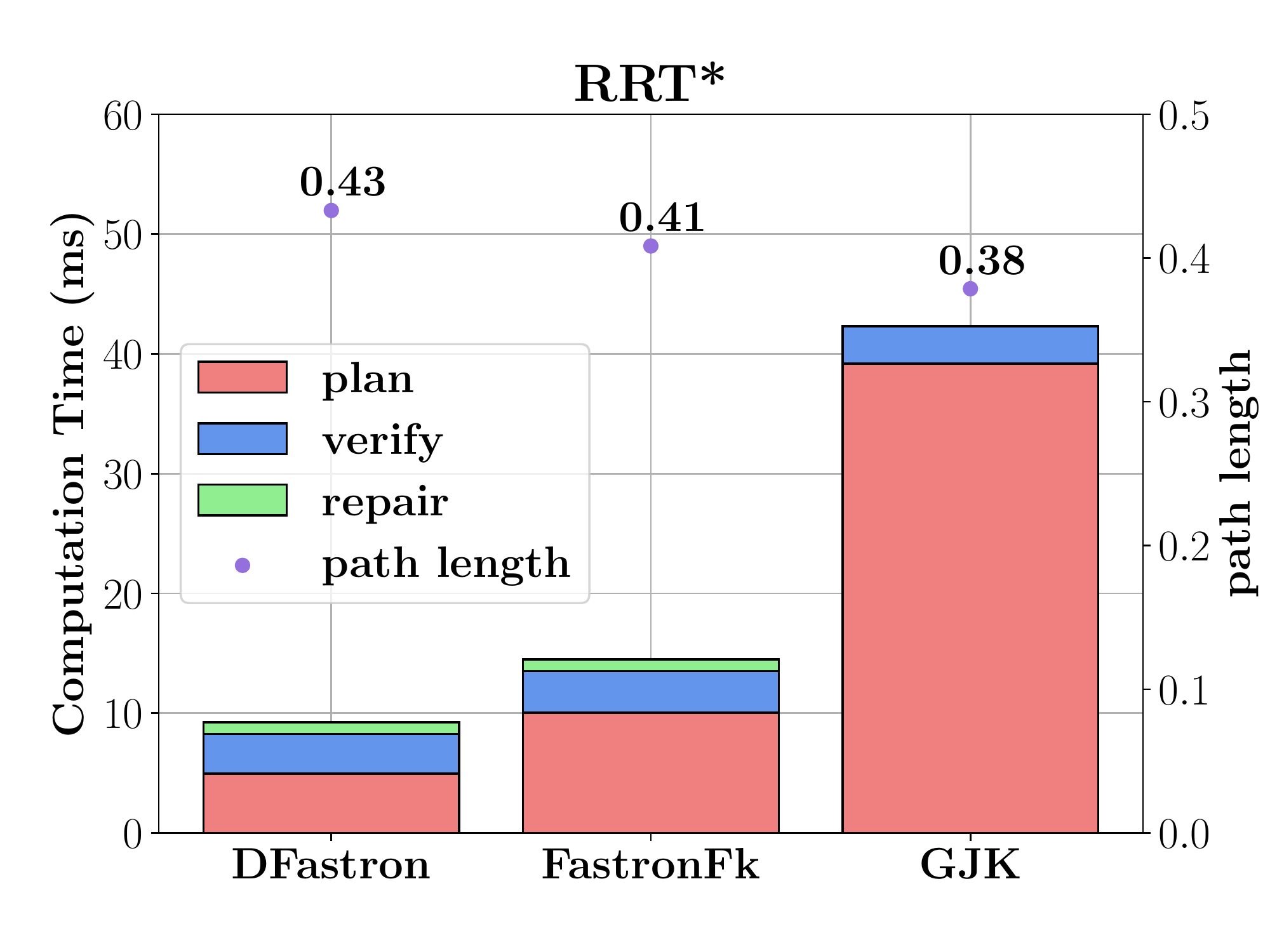}\\
       \includegraphics[width=.3\linewidth,trim={0cm 1cm 0cm 1cm},clip]{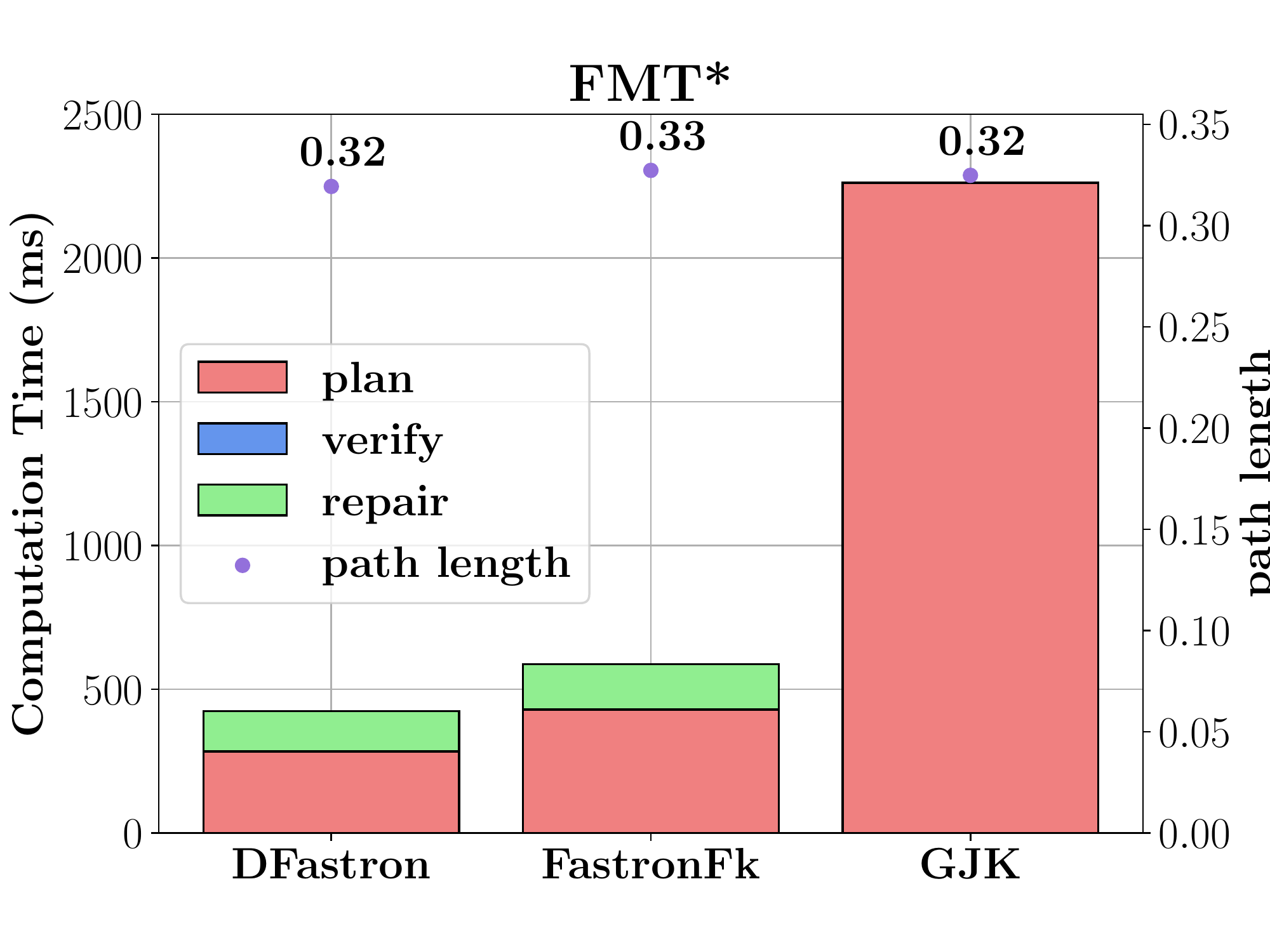}
       \includegraphics[width=.3\linewidth,trim={0cm 1cm 0cm 1cm},clip]{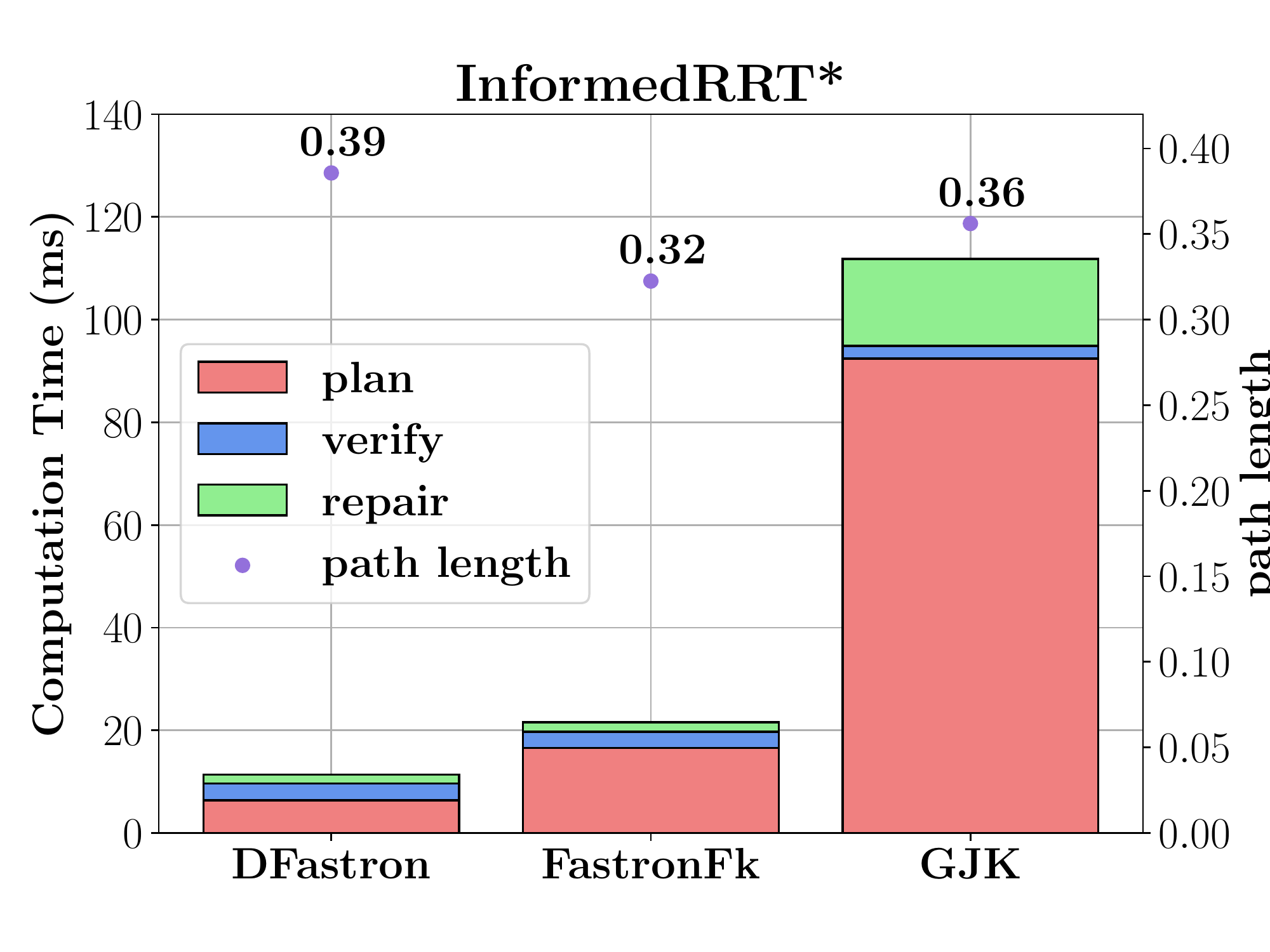}
       \includegraphics[width=.3\linewidth,trim={0cm 1cm 0cm 1cm},clip]{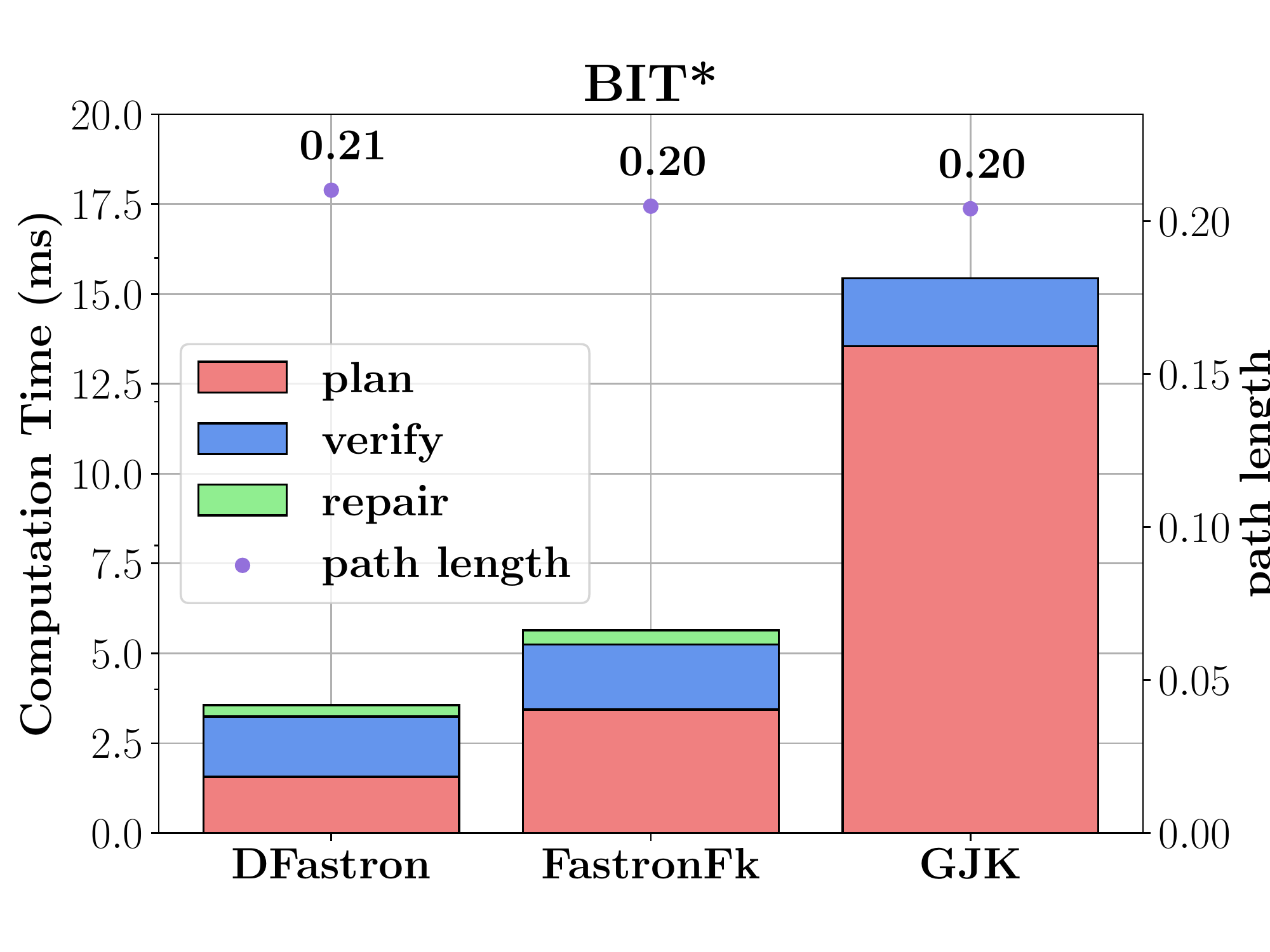}
  \caption{Plots of planning time for the following sampling based planning methods: RRT, SBL, RRT*, FMT*, Informed RRT*, and BIT*. The time spent planning, verifying the plan for integrity, and repairing if necessary are each shown red, blue, and green respectively. For the optimal planners that were stopped as soon as a solution was found, the average lengths of planned paths is shown by the purple dots and the right axis.}
  \label{fig:MotionPlanning} 
\end{figure*}

Lastly, we evaluate the performance of our new model in motion planning applications. We test D-Fastron, Fastron FK, and GJK in sampling-based motion planning using the Baxter robot. \updated{Prior work has shown GJK to be equivalent to or better than FCL for state-of-the-art motion planners \cite{FastronFK}.} We evaluate performance using the following planning algorithms from the Open Motion Planning Library in ROS: RRT, SBL, RRT*, FMT*, Informed RRT*, and BIT*. We generated an environment with 15 different randomly sampled objects for testing \updated{as shown in figure \ref{fig:Baxter}}. We then randomly selected the same start and goal points for all three models. Planning was repeated 20 times for each planner, and we averaged the results. RRT*, FMT*, Informed RRT*, and BIT* are optimal planners. Rather than letting them run until timeout, we stopped them as soon as they found a path. As all three collision checkers are approximate \updated{(our robot collision model is approximated with primitive shapes for GJK, making it also an approximate checker)}, we recorded the time spent planning, verifying the integrity of the plan, and repairing any mistakes. Verification and repair were done using FCL. More information on planning with proxy collision checkers can be found in \cite{Fastron Motion Planning}. Figure \ref{fig:MotionPlanning} shows the results of our motion planning tests. Average path lengths for the optimal planners are shown on the right axis, as we stopped these planners before they found an optimal solution. D-Fastron sees planning time improvements ranging from 1.38$\times$ to 1.89$\times$ faster over Fastron FK and from 2.89$\times$ to 9.82$\times$ faster over GJK in motion planning applications for complex environments depending on the planner used. BIT* and Informed RRT*, two state-of-the-art sampling-based planners showed 1.58$\times$ and 1.89$\times$ improvement respectively over Fastron FK.

 \section{Conclusion}
 In this paper, we present a novel method of configuration space decomposition to accelerate the Fastron FK proxy collision checker while maintaining comparable model correctness. We used K-Means clustering in the Forward Kinematic space to decompose the robot configuration space into multiple subspaces. Models trained in these subspaces can be queried quickly and scale better to more complex environments, increasing our collision checking performance compared to our previous work and other state-of-the-art methods. The decomposition is scalable, aware of the robot's kinematics, and does not rely on expert knowledge. As such, it is well suited to adaptations to other robots. It would be interesting to explore what other uses exist for this kind of C-space decomposition, possibly in other C-space models or other robot planning algorithms. When evaluating the performance of our models, we produce collision checking times 29$\times$ faster than geometric collision checkers like GJK and accelerate motion planning in high dimensional configurations spaces up to 9.8$\times$.  

 The ability to decompose the high-dimensional configuration spaces of robots for collision checking in an intelligent manner is timely given the rapid advances in parallel processing capabilities of GPUs and CPUs and will aid robots in becoming safer and more responsive in complex environments. 




\balance


\begin{thebibliography}{99}

\bibitem{MP Review}
 M. Elbanhawi and M. Simic, “Sampling-Based Robot Motion Planning: A Review,” \textit{IEEE Access}, vol. 2, pp. 56–77, 2014

 \bibitem{C-Space}
 T. Lozano-Perez, “Spatial Planning : A Configuration Space Approach,” \textit{IEEE Transactions on Computers}, vol. 32, no. 2, 1983
 
\updated{\bibitem{Collision Checking vs Nearest Neighbors}
M. Kleinbort, O. Salzman, and D. Halperin, "Collision Detection or Nearest-Neighbor Search? On the Computational Bottleneck in Sampling-based Motion Planning," \textit{Algorithmic Foundations of Robotics XII: Proceedings of the Twelfth Workshop on the Algorithmic Foundations of Robotics}, 2020, pp. 624-639}
 
\bibitem{Fastron Motion Planning}
N. Das and M. Yip, "Learning-Based Proxy Collision Detection for Robot Motion Planning Applications," in \textit{IEEE} Transactions on Robotics, vol. 36, no. 4, pp. 1096-1114, Aug. 2020, doi: 10.1109/TRO.2020.2974094.

\bibitem{FastronFK}
N. Das and M. C. Yip, "Forward Kinematics Kernel for Improved Proxy Collision Checking," in \textit{IEEE Robotics and Automation Letters}, vol. 5, no. 2, pp. 2349-2356, April 2020, doi: 10.1109/LRA.2020.2970645.

\bibitem{Pan SVM}
J. Pan and D. Manocha, “Efficient configuration space construction and optimization for motion planning,” \textit{Engineering}, vol. 1, no. 1, pp. 46–57, 2015.

\bibitem{Pan KNN}
J. Pan and D. Manocha, “Fast probabilistic collision checking for sampling-based motion planning using locality-sensitive hashing,” \textit{International Journal of Robotics Research}, vol. 35, no. 12, pp. 1477–1496, 2016.

\bibitem{Huh1}
J. Huh and D. D. Lee, “Learning high-dimensional mixture models for fast collision detection in rapidly-exploring random trees,” in \textit{2016 IEEE International Conference on Robotics and Automation, ICRA} 2016, Stockholm, Sweden, May 16-21, 2016, pp. 63–69.

\bibitem{Huh2}
J. Huh, B. Lee, and D. D. Lee, “Adaptive motion planning with high dimensional mixture models,” in \textit{2017 IEEE International Conference on Robotics and Automation}, ICRA 2017, Singapore, Singapore, May 29 - June 3, 2017, pp. 3740–3747.

\bibitem{Dimension Decomp}
Y. Han, W. Zhao, J. Pan, Z. Ye, R. Yi and Y. Liu, "A Configuration-Space Decomposition Scheme for Learning-based Collision Checking", in \textit{IEEE/RSJ International Conference on Intelligent Robots and Systems (IROS 20)}, pages 5678-5684, 2020

\bibitem{Octree Decomp}
T.H. Wong, G. Leach, and F Zambetta. "An adaptive octree grid for GPU-based collision detection of deformable objects.", Vis Comput 30, 729–738 (2014). https://doi.org/10.1007/s00371-014-0954-1

\updated{\bibitem{Quotient-Space}
A. Orthey, A. Escande, and E. Yoshida, "Quotient-Space Motion Planning", in \textit{2018 IEEE/RSJ International Conference on Intelligent Robots and Systems (IROS 18)}, pages 8089-8096, 2018}

\bibitem{DH}
C. R. A. Rocha, C. P. Tonetto, and A. Dias, “Robotics and Computer-Integrated Manufacturing A comparison between the Denavit Hartenberg and the screw-based methods used in kinematic modeling of robot manipulators,” \textit{Robotics and Computer Integrated Manufacturing}, vol. 27, no. 4, pp. 723–728, 2011. [Online]. Available: http://dx.doi.org/10.1016/j.rcim.2010.12.009

\bibitem{kmeans++}
D. Arthur and S. Vassilvitski "k-means++: The Advantages of Careful Seeding." \textit{Technical Report}. Stanford. (2006)

\bibitem{GJK}
 E. G. Gilbert, D. W. Johnson, and S. S. Keerthi, “A Fast Procedure for Computing the Distance Between Complex Objects in Three Dimensional Space,” \textit{IEEE Journal on Robotics and Automation}, vol. 4, no. 2, pp. 193–203, 1988
 
 \bibitem{BaxterFK}
 R.L. Williams II, “Baxter Humanoid Robot Kinematics”, \textit{Internet Publication}, https://www.ohio.edu/mechanical-faculty/williams/html/pdf/BaxterKinematics.pdf, April 2017.
 
 \bibitem{FCL}
 J. Pan, S. Chitta, and D. Manocha, “FCL: A general purpose library for collision and proximity queries,” \textit{Proceedings - IEEE International Conference on Robotics and Automation}, pp. 3859–3866, 2012.





\end{thebibliography}
\end{document}